# A Time-Series Foundation Model by Universal Delay Embedding

Zijian Wang[1,*], Peng Tao[1,*], Jifan Shi[3], Rui Bao[4], Rui Liu[5], Luonan Chen[1, 2,#]

[1] Key Laboratory of Systems Health Science of Zhejiang Province, School of Life Science, Hangzhou Institute for Advanced Study, University of Chinese Academy of Sciences, Hangzhou 310024, China

[2] School of Mathematical Sciences and School of AI, Shanghai Jiao Tong University, Shanghai 200240, China.

[3] Research Institute of Intelligent Complex Systems & CISOR, Fudan University, Shanghai, 200433, China.

[4] Frontiers Science Center for Deep Ocean Multispheres and Earth System, and Key Laboratory of Marine Chemistry Theory and Technology, Ministry of Education/ Institute for Advanced Ocean Studies, and Innovation Center for Ocean Carbon Neutrality, Ocean University of China, Qingdao, China, Qingdao,266000

[5] School of Mathematics, South China University of Technology, Guangzhou 510640, China.

[*] Contribution equally

[#] To whom correspondence should be addressed. Luonan Chen, E-mail: lnchen@sjtu.edu.cn

**Abstract**

This study introduces Universal Delay Embedding (UDE), a pretrained foundation model designed to revolutionize time-series forecasting through principled integration of delay embedding representation and Koopman operator prediction. Leveraging Takens' embedding theorem, UDE as a dynamical representation of observed data constructs two-dimensional subspace patches from Hankel matrices, theoretically preserving dynamical and topological properties of underlying dynamical systems. Such patches are viewed as images, which can be efficiently processed by exploiting advanced deep learning technologies. Computationally, these patches further serve as tokens for learning a self-attention encoder, thus enabling accurate prediction of nonlinear time-series by a finite-dimensional Koopman operator in a linear manner in a latent space. Extensive evaluations across various benchmarks and real-world climate datasets demonstrate over 20% average reduction in mean squared error versus state-of-the-art foundation models, alongside superior generalization in fine-tuning scenarios. In particular, the learned dynamical representations and Koopman operator prediction forms from the patches exhibit exceptional interpretability, with consistent identification of topologically informative subspaces and robust encoding of domain-invariant dynamics, establishing UDE as a scalable, interpretable framework for universal time-series modeling and forecasting with broad scientific and industrial applicability.

# Introduction

Time series forecasting (TSF) aims to predict future values from historically observed, temporally ordered data points[1,2]. As a cornerstone of machine learning and statistics, TSF has critical applications across various domains, including finance[3,4], environmental science[5], epidemiology[6], and industry[7,8]. Early baselines were established by classical statistical models, such as ARIMA[9,10], VAR[11], and GARCH[12], as well as machine learning algorithms, including support vector regression[13] and random forests[14]. Subsequent advancements in recurrent neural network units (e.g., LSTM and GRU[15–17]), temporal convolutional networks (TCN)[18], and Transformer-based architectures[19] have further improved forecasting accuracy. However, these methods are typically trained and evaluated on datasets confined to specific domains. Consequently, their predictive performance often degrades significantly when applied to other domains—a phenomenon commonly referred to as the cross-domain generalization problem[20,21,22].

To enhance cross-domain generalization, time-series foundation models leveraging universal Transformer architectures have emerged[23–25]. These frameworks learn transferable temporal representations from diverse datasets, enabling adaptation without exhaustive retraining[23,26–30]. Existing time-series foundation models generally follow two main paradigms: pretraining from scratch and LLM-based. In the first approach, models are pretrained on time series from multiple domains to capture domain-invariant temporal patterns. Forecasting is then performed either via direct state projection or through autoregressive decoding (e.g., next-token prediction)[26,28,29]. In contrast, the LLM-based approach leverages the powerful sequence modeling capabilities of pretrained language models by embedding time series into a tokenized space compatible with LLM inputs. This alignment is typically achieved via two strategies: (i) converting time series into pseudo-natural language representations and fine-tuning only a subset of LLM parameters while freezing the core Transformer blocks[23,31], or (ii) embedding time series data to LLM prompts[25,32]. However, due to the fundamental differences between continuous-valued time series (typically dynamical systems) and discrete-valued natural language sequences (generally probabilistic models), the efficacy of LLMs in time series forecasting remains uncertain[33]. This underscores the importance of building foundation models based on dynamical systems pretrained from scratch on time series data, which can better capture domain-specific dynamics and temporal invariants. Consequently, such models constitute the central focus of the present study.

Nevertheless, despite their superior generalization capabilities, time-series foundation models typically underperform compared to state-of-the-art (SOTA) domain-specific deep-learning approaches (e.g., PatchTST[34] and iTransformer[35]) on individual benchmarks, and their internal mechanisms are considerably more opaque. Our systematic analysis of existing time-series foundation models indicates that this underperformance and limited interpretability primarily originate from three factors: (i) the exclusive reliance on fixed temporal tokenization schemes (e.g., daily, weekly, yearly, seasonal) to extract latent temporal patterns, (ii) the excessive architectural complexity of these models, which is often introduced without theoretical justification, and (iii) many of them mainly adopt static representation of high-dimensional data rather than dynamical features. Consequently, there is an urgent need to develop novel theoretical frameworks and model architectures that explicitly address these deficiencies[36–38].

According to dynamical systems theory, time series are continuous-valued sequences which are typically regarded as observations from an underlying dynamical system evolving over time. Although the original system may reside in a high-dimensional space, its effective dynamics are typically governed by a lower-dimensional manifold embedded within that space. Specifically, according to Takens' embedding theorem[20], it is possible to reconstruct the whole system's attractor and preserve its topological properties using time-delay coordinates from a single observed variable. This theoretical foundation has inspired a machine learning framework that incorporates time-delay embeddings to capture both short and long term dependencies. Furthermore, under certain formulations, time-delay systems can be mapped into high-dimensional or even infinite-dimensional functional spaces (e.g., Hilbert spaces), enabling the approximation of complex dynamics with great expressive power. Inspired by the time-delay theory and generalized embedding theory, the spatiotemporal information (STI)[39–41] transformation framework is able to convert high-dimensional spatial information (i.e., high-dimensional system states) into the temporal information (i.e., one-dimensional time sequences) of latent or target variables, and vice versa, thus providing a way for dynamical representation of high-dimensional data. Several methods have been proposed for time series prediction by solving STI equations, including RDE[41], ARNN[39], MT-GPRM[42], Delayformer[43], HoGRC[44], FRMM[45]. In addition to STI, methods grounded in dynamical systems theory (e.g., neural ODEs[46], PINNs[47], and Koopman[48]) emphasize modeling the intrinsic dynamics underlying observed data rather than solely predicting future states. This approach enhances both interpretability and predictive accuracy.

Inspired by dynamical systems theory, we propose Universal Delay Embedding (UDE), a pretrained foundation model for time-series forecasting to alleviate those deficiencies. In contrast to existing foundation models, UDE transforms the raw data of each variable into a delay-embedded (dynamical) representation of whole original system in a latent space one by one and further tokenizes each of them via subspace projections formed as patches, thereby preserving dynamical features that conventional tokenization schemes neglect. These patches can be processed as images, thus not only capturing implicit static/dynamical patterns but also improving the efficiency by exploiting the advanced deep learning technologies (e.g. Vision Transformers (ViT)[51]). Furthermore, drawing upon Koopman operator theory, UDE adopts a simpler encoder-only architecture by lifting such a latent space to a higher dimension for prediction, which renders the forecasting task linear. These two innovations of delay-embedded representation and Koopman operator prediction, both grounded in rigorous dynamical-systems principles, endow UDE with superior interpretability and markedly enhanced predictive performance relative to current foundation models.

Evaluated on benchmark and real-world climate datasets, UDE achieves superior accuracy, robustness to data irregularities (e.g., nonstationarity, noise), and strong zero/few-shot transfer. Analyses confirm its capacity to differentiate cross-domain inputs while consistently mapping same-domain data to similar latent distributions. Our work substantiates the value of forecasting inspired by dynamical systems, and pioneers the integration of delay embeddings and Koopman operator prediction with a pretraining model, opening avenues for scientific and industrial innovation.

## Results
**Time delay patching from each observed variable revealing local subspace and temporal structures**

For a multivariate time series $x^t = [x_1^t, \dots, x_n^t]' \in \mathbb{R}^n$ observed over a sliding window $t = 1, 2, \dots, T$ from an underlying $n$-dimensional system with each variable as $x_i^t \in \mathbb{R}$, the goal of TSF is to predict future states $x^t, t > T$ by learning the latent temporal structure of the observed data. In most deep learning frameworks for TSF, a channel-independence approach is commonly adopted due to its high generalizability and scalability. However, this approach often fails to capture the inter-variable dependencies and global structure of the underlying system. To overcome this limitation, we proposed time delay patching (TDP) by constructing a discrete time-delay embedding system under Takens' framework (Supplementary Note 4), parameterized by an embedding dimension $m$ and a delay step $\tau$ from a univariate time series of length $T$. Each delay embedding vector from one variable $x_i$ is defined as:

$$y_i^t = \left[x_i^{t-(m-1)\tau}, \ldots, x_i^{t-2\tau}, x_i^{t-\tau}, x_i^t\right]' \in \mathbb{R}^m \tag{1}$$

which captures the local state of the system from a single observation channel. Within the time span of $T$, we extract $L = T - (m-1)\tau$ such delay embedding vectors, collectively reconstructing the system's trajectory or states in phase space from this observed scalar time series $x_i$. In other words, it can theoretically be proven that there generically exists a smooth map $\phi_i$ such that for $t = (m-1)\tau + 1, (m-1)\tau + 2, \ldots, (m-1)\tau + T$

$$y_i^t = \phi_i(x^t),$$

provided that $m > 2d$ where $d$ is the box-counting dimension of the attractor and is usually small (Supplementary Note 4). Hence, such a delay-embedded representation $y_i^t$ can be viewed as a dynamical representation[39–41] of the original state $x^t$ in contrast to the static representation used widely in deep learning (see Method and the following section). Note that there generically exists the reversed $\phi_i^{-1}$ such as $x^t = \phi_i^{-1}(y_i^t)$, and thus such a dynamical representation can also be viewed as a universal approximation framework or UDE, which is theoretically ensured by STI[39-45].

Specifically, we stack the $L$ delay vectors into a delay-embedding matrix (i.e., a generalized Hankel matrix) defined as:

$$H_i = \begin{bmatrix} x_i^1 & x_i^2 & \cdots & x_i^m \\ x_i^2 & x_i^3 & \cdots & x_i^{m+1} \\ \vdots & \vdots & \ddots & \vdots \\ x_i^L & x_i^{L+1} & \cdots & x_i^{m+L-1} \end{bmatrix}, \tag{2}$$

where the delay step $\tau$ is set to 1. Under Takens' theorem, this embedding space from a single variable $x_i$ is generically homeomorphic to the attractor of original high-dimensional system, thereby preserving its topological and dynamical structure. Thus, the time delay embedding matrix provides a topology-aware (spatial) representation of the temporal evolution, forming the basis for subsequent patch-level tokenization and forecasting. Note that $H_i$ is considered as one form of $y_i$ in Eq. (1).

In practice, the delay step $\tau$ is typically set to 1 to fully exploit the native temporal resolution. This choice yields densely sampled delay vectors comprising consecutive observations[49]. When $\tau = 1$, the time-delay embedding matrix reduces to a strict Hankel matrix, which is advantageous for subspace analysis, although alternative criteria[50], such as mutual information or autocorrelation, can be employed to estimate an optimal $\tau$.

To computationally facilitate spatiotemporal tokenization analogous to the 2-D patching mechanism in ViT[51], we partition the time-delay embedding matrix $H_i \in \mathbb{R}^{L \times m}$ into non-overlapping 2-D rectangular patches. Let $U = \lfloor \frac{L}{p} \rfloor$ and $V = \lfloor \frac{m}{q} \rfloor$ be the number of patch rows and columns, respectively, where $\lfloor \rfloor$ denotes the floor function. The total number of patches is $U \times V$. Each patch $\boldsymbol{P}_i^{(j)} \in \mathbb{R}^{p \times q}$, with index $j = (u-1)V + v$ for $u = 1, \ldots, U$ and $v = 1, \ldots, V$ can be defined as:

$$\boldsymbol{P}_i^{(j)} = \boldsymbol{H}_i[(u-1)p + 1 : up, (v-1)q + 1 : vq], \tag{3}$$

where the notation $a:b$ denotes the range of rows or columns from $a$ to $b$ (inclusive). $\boldsymbol{P}_i^{(j)}$ spans $p$ consecutive time steps (rows) and $q$ delayed coordinates (columns) in the time delay system, capturing a local geometric segment of the time-delay embedding space and preserving subspace dynamics along both temporal and embedding axes. Each patch can therefore be interpreted as a projection of the full time-delay trajectory onto a local subspace that encodes short-term evolution and the local topology of the system's phase space (see Methods for details).

Fig. 1 illustrates the structural fidelity of TDP across four distinct dynamic regimes: periodic, random-walk, chaotic, and sparse-pulse patterns. In each case, TDP retains fine-grained geometric patterns and subspace-aware representations, thereby significantly outperforming conventional 1-D slicing methods.

Periodic sequences (Fig. 1a–c) exhibit strong structural regularity in delay-embedding space. While conventional 1-D slicing (Fig. 1c) partitions the signal along the temporal axis and partially retains short-range periodicity, it fragments the global cycle and disrupts inter-patch consistency. In contrast, TDP (Fig. 1b) transforms the signal into a high-dimensional delay-embedding trajectory, and extracts localized 2-D patches. TDP preserves the underlying periodic geometry and local topology, allowing the model to perceive and exploit repeating patterns consistently across the sequence.

Random-walk sequences (Fig. 1d–f), characterized by non-stationarity and path dependence, present challenges for conventional slicing (Fig. 1f), which yields fragmented patches dominated by noise and lacking coherent structure. TDP (Fig. 1e) reconstructs the delay trajectory and reveals smooth transitions within patch intensities—often forming gradient-like patterns across rows or diagonals. These gradients encode the local drift direction of the walk, offering insight into trend orientation (upward, downward, or reversal). By capturing delayed correlations, each patch reflects not only instantaneous fluctuations but also structured movement within the time-delay embedding space. This provides a topology-informed representation better suited for modeling non-stationary and drift-driven behaviors.

Chaotic sequences from the Lorenz attractor (Fig. 1g–i) are globally unpredictable but locally structured. Conventional slicing (Fig. 1i) fails to uncover these patterns, producing disordered segments with minimal dynamical coherence. In contrast, TDP (Fig. 1h) exploits the theoretical guarantee that delay embeddings are topologically equivalent to the original system's attractor under Takens' theorem. Each patch thus acts as a localized projection of the chaotic trajectory that retains the attractor's topology, enabling the model to infer fine-grained dynamical flow patterns and capture complexity even under chaos.

Sparse-pulse sequences (Fig. 1j–l) highlight another advantage of TDP. Traditional slicing (Fig. 1l) often cuts through pulse events or dilutes them with low-value segments, resulting in information loss. TDP (Fig. 1k), however, embeds each pulse alongside its surrounding temporal context, forming distinct 2-D patterns. These patches encode both the presence and timing of pulse events, allowing the model to infer inter-pulse relationships in a spatially structured manner. This is particularly beneficial for capturing temporal sparsity and event localization.

In summary, TDP extracts local 2-D subspace projections from delay-embedding systems in which each patch retains the topological and dynamical properties of the original signal from each observed variable. In particular, we can process such a matrix form as an image by exploiting the advanced deep-learning methodologies (e.g., ViT), thus not only capturing implicit dynamical patterns but also improving computational efficiency greatly. For chaotic systems, this ensures that local patches remain homeomorphic to the original attractor. For drift- or pulse-driven inputs, patches reveal smooth directional trends or spatial event signatures, respectively. By preserving the geometric structure of time-delay trajectories, TDP enables dynamics-aware tokenization or dynamical representation that markedly enhances model interpretability and performance across diverse temporal patterns.

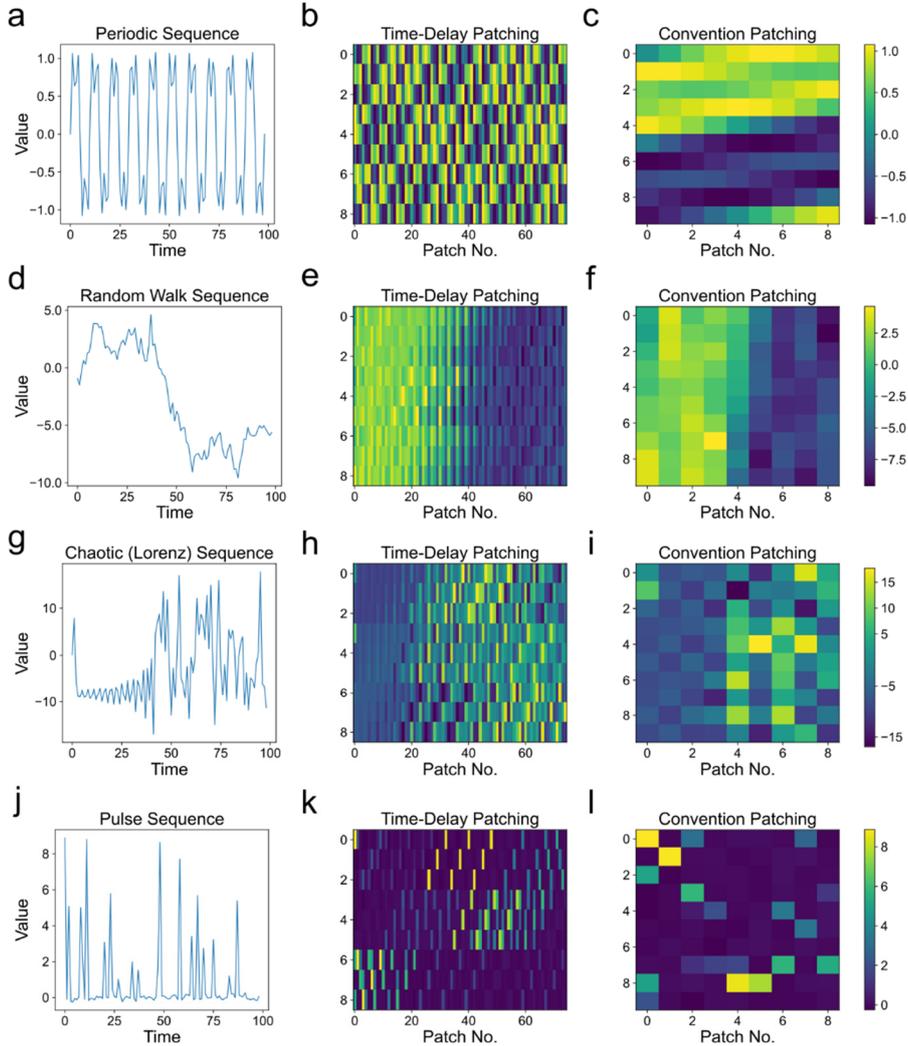

**Fig. 1. Structural fidelity of time delay patching (TDP) across time series with distinct dynamical properties.** Raw time-series signals exhibit four distinct types of dynamics: **(a)** periodic, **(d)**. random-walk, **(g)** chaotic (Lorenz), and **(j)** sparse-pulse patterns. The second column (**b**, **e**, **h**, **k**) presents flattened dynamical representations of 2-D patches extracted from the corresponding delay-embedding matrices (embedding dimension $m = 9$, delay step $\tau = 1$). Each non-

overlapping patch spans both the temporal and delay axes, capturing a local trajectory segment in the time-delay embedding space. The third column (**c, f, i, l**) shows results from conventional 1-D patching based on uniformly slicing the raw sequence along the temporal axis. Compared to the loss of structural features in conventional patching, TDP preserves fine-grained geometric patterns, yielding subspace-aware dynamical representations of the underlying dynamics.

**Framework of universal delay embedding (UDE) as a dynamical representation**
We first theoretically derive the UDE as a foundation model of time-series prediction from a mathematical viewpoint and then provide the algorithmic framework from a computational viewpoint.

After constructing the delay embedding vector $y_i^t$ in Eq. (1), we then map it to a higher-dimensional latent space using a learnable network, such as a Transformer, i.e., lifting the lower-dimensional representation $y_i^t$ to a higher-dimensional dynamical representation $z_i^t \in \mathbb{R}^M$,

$$z_i^t = f(y_i^t) \qquad (4)$$

where $f$ is a nonlinear function implemented by a Transformer model across all $i$. Thus, this lifted higher-dimensional representation $z_i^t = [z_i^t, z_i^{t+1}, \ldots, z_i^{t+M-1}]' \in \mathbb{R}^M$ is considered to transform nonlinear dynamics $y_i(t)$ approximately to a linear space $z_i^t = f(y_i^t) = f(\phi_i(x^t))$ with a higher-dimension $M$ (> m) by further training a common linear layer $g(\cdot)$. Specifically, to model the temporal evolution in the high-dimensional space, we approximate the action of the Koopman operator $\mathcal{K}$ by a trainable prediction linear layer $g(\cdot)$ common to all $i$ for $h$-step prediction:

$$z_i^{t+h} = g(z_i^t) \qquad (5)$$

where $\tau = 1$ for simplicity. Although outputs are channel-specific, we tie the prediction head across channels. After delay embedding and a shared encoder, all variables are lifted to a common Koopman-like latent space with a higher dimension $M$. The existence of a smooth decoding map $\Phi(\cdot)$ is proved in Supplementary Note 4 based on Eq. (4) and the definition of $\phi_i$, such that the whole original system observation $x^t$ can be reconstructed from the dynamical representation of one variable $x_i$ or its sequence: $x^{t+h} = \Phi_i(z_i^{t+h})$, where $\Phi_i = (f\phi_i)^{-1}$ for each $i$. Thus, from Eqs. (4)-(5) and $\Phi_i$, we can theoretically predict $\hat{x}^{t+h} = \Phi_i\left(g(z_i^t)\right) = \Phi_i\left(g\left(f(y_i^t)\right)\right) = \Phi_i\left(g(f(\phi_i(x^t)))\right)$ in a closed form. In this work, to build a general or foundation model, we predict $x_i$ separately from $z_i$ but by a common $\Phi$ of all $i$, i.e.

$$\hat{x}_i^{t+h} = \Phi(z_i^{t+h}). \qquad (6)$$

Thus, from Eqs. (1) and (4)-(6), theoretically, we have the $h$-step prediction $\hat{x}_i^{t+h}$ form from one variable $y_i$ or $x_i$

$$\hat{x}_i^{t+h} = \Phi\left(g\left(f(y_i^t)\right)\right) = \Phi\left(g(f(\phi_i(x^t)))\right).$$

Next, we show how to construct $f, g, \Phi$ of Eq. (4)-(6) computationally from one variable $x_i$ or actually $y_i$ by proposing the UDE framework as a foundation model for time-series forecasting (Fig. 2). This framework operationalizes the theoretical foundations of time-delay embedding and manifold-preserving dynamics into a scalable Transformer-based model for universal time-series forecasting.

As shown in Fig. 2a–b, UDE or $(f, g, \Phi)$ in Eqs. (4)-(6) begins by applying time-delay embedding independently to each input variable, converting scalar sequences into Hankel matrices that encode the system's latent phase-space evolution as a dynamical representation, different from the static representation used currently in deep learning (see Fig.2e-f). Under Takens' theorem, the set of Hankel matrices $\mathcal{H} = \{H_1, \ldots, H_c\}$ derived from all $c$ variables are pairwise homeomorphic to projections of the same underlying dynamical attractor $\mathcal{A}$. This topological equivalence allows the use of a single shared mapping $\Phi: \mathbb{R}^{L \times m} \to \mathbb{R}^h$, applied identically to each $H_i$ or $y_i$ to produce its $h$-step forecast, i.e., $\hat{x}_i^{t+h} = \Phi\left(g\left(f(y_i^t)\right)\right)$ for all $i$ to learn a common attractor and Koopman operator, which exhibits cross-domain consistency in topological invariants and spectral structure. Therefore, each variable can be processed independently passed through the same model parameters for $(f, g, \Phi)$ at a time $t$, to obtain its forecast at $t + h$.

These $H_i$ are partitioned into non-overlapping 2-D patches along both the time and delay axes by TDP. Each patch defines a local subspace, namely, a short segment of the delay trajectory, that captures fine-grained temporal geometry before being flattened and projected into a token. Similar to the imaging processing, these tokens are processed by a shared encoder-only (trainable) Transformer equipped with multi-head self-attention, enabling global interaction while preserving subspace locality. The resulting or lifted latent embedding $z_i$ by $f$ is the representation vector with a much higher dimension, which are then aggregated and passed through a shared (trainable) MLP prediction head, yielding multivariate forecasts across diverse domains as an approximation of the Koopman operator, i.e. a linear map $g$. Furthermore, the predicted $z_i$ is used to reconstruct the prediction of $x_i$ for each $i$ by a common map $\Phi$. Clearly, we can train the common neural network $(f, g, \Phi)$ for each observed $i$, which learns all the inherent information of the original system $x$ across all $i$. Therefore, it can make the prediction of $x$ as a foundation model by forecasting $x_i$ one by one. This modular structure (Fig. 2c) enables UDE or $(f, g, \Phi)$ to be pretrained once on large-scale, heterogeneous time-series data and deployed directly on new domains in a zero-shot or few-shot fashion, with optional fine-tuning of the prediction head. Note that we combine $(g, \Phi)$ as a linear map in this work (see Fig.2), which is able to make accurate predictions although a nonlinear map can be adopted.

In addition to the theoretical foundation, critically UDE's architecture also promotes interpretability. As shown in Fig. 2d, the learned token space exhibits strong topological organization. By computing pairwise 2-Wasserstein distances (Supplementary Note 3) between the persistence diagrams of token, UDE identifies coherent token clusters that correspond to dominant dynamical patterns. These clusters are consistently attended to across encoder layers, acting as "dynamical anchors" for global modeling. This attention-based prioritization of topologically informative subspaces highlights UDE's ability to learn semantically meaningful and structurally consistent representations of temporal dynamics.

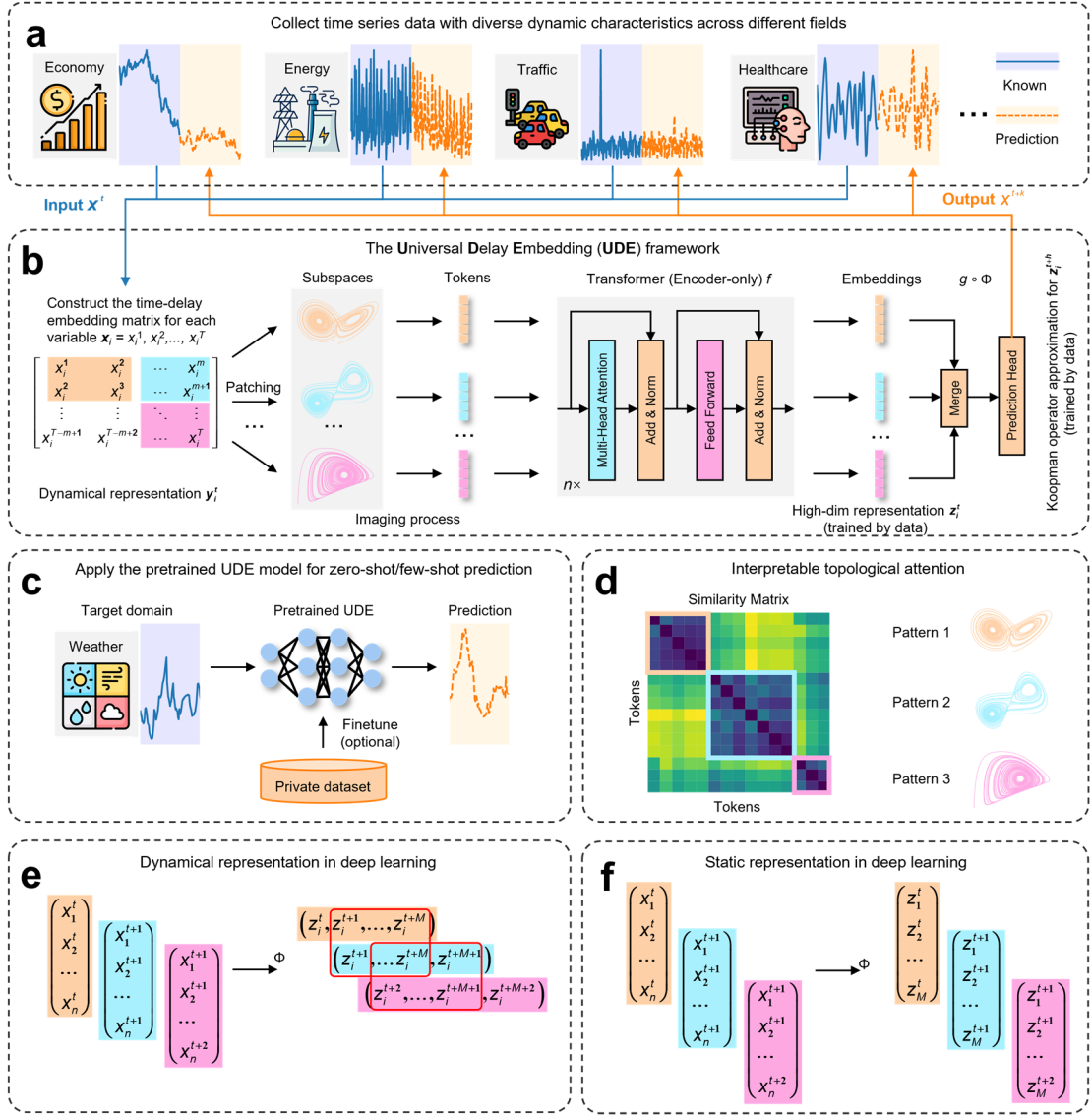

**Fig. 2 Universal Delay Embedding (UDE) structure as a foundation model for multivariate time-series forecasting.** (**a**) Multi-domain training data spanning economy, energy, traffic, and healthcare etc. (**b**) Time-delay embedding matrices are constructed for each observed variable $x_i$ (we can have same procedure for any $i$) as dynamical representations ($y_i^t$ or $z_i^t$) of the original high-dimensional data, partitioned into sub-space patches $P_i^{(j)}$, and tokenized before being fed into an encoder-only (trainable) Transformer $f$; the delay embedded space/vector is actually transformed to a much higher dimension of the latent vector "Embedding" ($z_i^t$), which is further used to make the forecast ($z_i^{t+h}$ or $x_i^{t+h}$ by the (trainable) MLP prediction head as an approximation of $(g, \Phi)$. (**c**) Zero-/few-shot deployment for target domain (e.g., weather) forecasting, with optional fine-tuning on private data. (**d**) Interpretable topological-attention map revealing distinct canonical dynamical patterns. (**e**) Dynamical representation of Eq. (4) or $x^t = \Phi_i(z_i^t)$ means that each state vector $x^t$ at time instant $t$ is represented by a latent sequence (vector) $z_i^t = (z_i^t, z_i^{t+1}, \ldots, z_i^{t+M-1})$ with the same scalar variable $z_i$ but at different time instants $t, t+1, \ldots, t+M-1$, preserving trajectory geometry and dynamics. (**f**) Static representation means that each state vector $x^t$ at time $t$ is represented by a latent vector $\mathbf{z}^t = (z_1^t, z_2^t, \ldots, z_M^t)$ with different

variables $(z_1, z_2, ..., z_M)$ but at the same time $t$, thus lacking explicit dynamical structure.

**UDE learns structured subspace tokens for dynamical representation**

To further investigate how UDE organizes its dynamical representations, we first pretrained a small-scale (6 layers, 8 heads) UDE on UTSD-1B[28] data and then analyzed the learned subspace tokens and attention dynamics with a 512-step ETTh1 sequence input (Fig. 3a). The time-delay embedding of this sequence (Fig. 3b) reveals a globally coherent geometric trajectory, which UDE decomposes into localized patches via TDP. Each patch captures a distinct segment of the system's evolution in phase space and forms an input token for the Transformer encoder.

We computed pairwise 2-Wasserstein distances between tokens based on their persistence diagrams[52–55] (Fig. 3c). These distances quantify differences in topological structures, such as the number and lifespan of connected components and loops, rather than the raw value similarity. This analysis reveals the emergence of distinct token clusters, suggesting that UDE implicitly organizes its token space based on underlying dynamical patterns instead of positional or temporal order. The complete pairwise 2-Wasserstein distance matrix for all 81 tokens (Supplementary Fig. 1), together with the detailed calculation procedure, is provided in Supplementary Note 2. Fig. 3d visualizes each representative subspace from the four clusters, tokens 1, 17, 6 and 8. Pattern 1 is markedly more representative than the remaining three. This structural coherence is reflected in the cross-layer attention maps (Fig. 3e), where bright vertical stripes indicate that a small subset of tokens consistently attract attention from all others across encoder layers and heads. These attention-dominant tokens act as dynamical anchors, summarizing globally relevant behaviors of the system. The token frequency histogram (Fig. 3f) further supports this finding: tokens of No. 19, 1, 11, and 3 appear most frequently as high attention tokens, which are defined as tokens with attention scores exceeding the mean by two standard deviations in one attention head. These four high-attention tokens correspond exactly to the highlighted columns in Fig. 3e and are members of Pattern 1, confirming that UDE not only identifies structurally representative subspaces but also prioritizes them as central components of its modeling process.

Collectively, these results demonstrate that a pre-trained UDE learns to extract and organize a sparse set of geometry-aware subspace tokens that reflect the intrinsic dynamics of the underlying system. Through consistent attention allocation and topologically meaningful token formation, UDE constructs a representation space that is both interpretable and dynamically expressive.

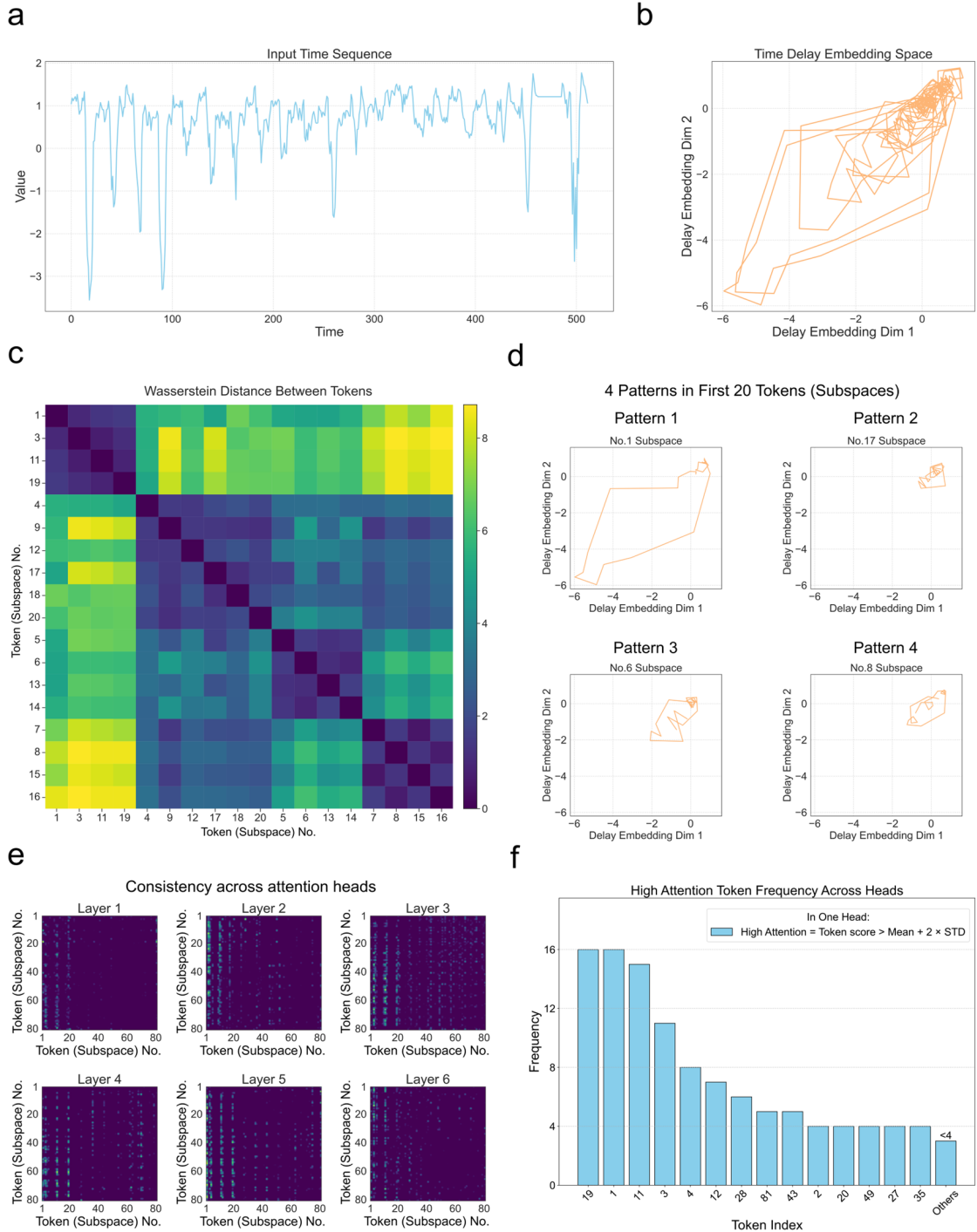

**Fig. 3. UDE identifies topologically informative subspaces through attention consistency. (a)** Input sequence (ETTh1, length 512) used for subspace analysis. **(b)** Visualization of the full delay embedding system in two selected dimensions, illustrating the global geometry of the reconstructed trajectory. **(c)** Wasserstein distance matrix between the first 20 tokens (subspaces), measuring pairwise geometric similarity in the delay embedding space. Clear block structures suggest four groups of topologically similar tokens. **(d)** Visualization of four representative token (No. 1, 17, 6, 8) of the four patterns calculated in **(c)**, showing distinct local trajectory patterns that recur across the delay embedding manifold. These patterns correspond to key dynamical modes extracted by the model. **(e)** Attention maps across six transformer layers. Each heatmap displays attention from all

81 tokens (queries, rows) to others (keys, columns) in one attention head. High-consistency vertical stripes indicate a small set of tokens that consistently attract attention across heads and layers. **(f)** Frequency histogram of high-attention tokens. For each encoder head in each layer, tokens with attention scores exceeding the mean by two standard deviations are counted. Token 19, 1, 11, 3 appear most frequently, indicating they serve as globally informative, topologically representative summary tokens, which indicates that the four tokens represent the same dynamic properties. As shown in (**c-d**), these tokens form Pattern 1, a group of subspaces with highly similar geometry. This pattern reflects the most representative dynamical structure of the full time-delay embedding system, and its recurrence across layers and heads suggests that UDE consistently anchors its representations on this dominant trajectory manifold.

**Zero-shot performance of pretrained UDE across benchmarks**

To evaluate the forecasting capabilities of UDE, we pre-trained three model variants of increasing size—UDE-Small (6 layers, 8 heads), UDE-Medium (8 layers, 12 heads), and UDE-Large (10 layers, 16 heads)—on the large-scale USTD-12B[28] dataset (Supplementary Note 1). To enhance computational efficiency without sacrificing representation quality, we integrated a token-pooling module commonly employed in vision tasks. Specifically, we applied average pooling over the patch tokens using a kernel size and stride of 30.

We assessed zero-shot forecasting performance on seven widely used benchmarks: ETTh1, ETTh2, ETTm1, ETTm2, Weather, Traffic, and Electricity. All models were evaluated under a 96-step prediction horizon without task-specific fine-tuning. As shown in Table 1, UDE demonstrates strong generalization across all datasets compared to mainstream time-series foundation models. Notably, UDE-Small achieves the best overall average mean squared error (MSE) of 0.264, outperforming all compared models, including those trained on far larger datasets or with far more parameters, such as Timer-16B (0.343), Moirai-M (0.326), and Chronos-S20 (0.326). Even the lightweight UDE-Small variant surpasses models with billions of parameters on key datasets such as ETTh2 (0.197), ETTm2 (0.140), and Weather (0.179), while maintaining highly competitive performance on Traffic (0.307) and Electricity (0.151). Compared with the best-performing baseline, Timer-16B, UDE achieves lower error on five of the seven datasets (23% average reduction in MSE). This demonstrates the strong inductive bias and transferability provided by UDE's subspace-aware delay tokenization and its ability to learn dynamics-consistent patch representations during pre-training.

Collectively, these results illustrate that UDE offers both parameter efficiency and robust zero-shot generalization across diverse temporal scenarios, including electricity-load prediction, weather-pattern forecasting, and complex, market-like fluctuations.

**Table 1 Zero-shot forecasting performance across seven benchmark datasets (96-step MSE).**
Results are reported for UDE variants and existing time series foundation models. **Bold** indicates the best result, and *underlined* indicates the second-best for each dataset.

| | Zero-shot (MSE) | | | | | | | |
|---|---|---|---|---|---|---|---|---|
| | ETTh1 | ETTh2 | ETTm1 | ETTm2 | Weather | Traffic | Electricity | Average |
| **UDE-small** | 0.376 | **0.197** | 0.498 | **0.140** | **0.179** | **0.307** | 0.151 | **0.264** |
| **UDE-medium** | 0.378 | 0.215 | 0.476 | *0.141* | 0.183 | *0.308* | 0.153 | *0.265* |
| **UDE-large** | *0.371* | *0.208* | 0.483 | 0.144 | 0.186 | 0.311 | 0.155 | *0.265* |
| TIMER-1B[28] | 0.438 | 0.314 | 0.690 | 0.213 | *0.181* | 0.458 | 0.192 | 0.355 |
| TIMER-16B[28] | **0.364** | 0.294 | 0.766 | 0.234 | 0.203 | *0.399* | **0.139** | 0.343 |
| TIMER-28B[28] | 0.393 | 0.308 | 0.420 | 0.247 | 0.243 | 0.414 | *0.147* | 0.310 |
| MOIRAI-S[29] | 0.400 | 0.341 | 0.448 | 0.300 | 0.242 | 0.616 | 0.233 | 0.369 |
| MOIRAI-M[29] | 0.434 | 0.345 | *0.381* | 0.272 | 0.238 | 0.425 | 0.188 | 0.326 |
| MOIRAI-L[29] | 0.510 | 0.354 | 0.390 | 0.276 | 0.259 | 0.399 | 0.188 | 0.339 |
| MOMENT[27] | 0.674 | 0.330 | 0.670 | 0.257 | 0.255 | 1.293 | 0.744 | 0.603 |
| TIMESFM[26] | 0.414 | 0.318 | **0.354** | 0.201 | - | - | - | - |
| CHRONOS-S1[23] | 0.571 | 0.423 | 0.632 | 0.272 | - | - | - | - |
| CHRONOS-S20[23] | 0.454 | 0.326 | 0.451 | 0.190 | - | - | - | - |

**Data-efficient finetuning of UDE**

We next examine whether UDE retains its advantages in few-shot regimes, a critical requirement for real-world deployments where labelled data are scarce or costly. We fine-tuned the pre-trained UDE-Small model (introduced in Table 1) on two representative datasets, i.e., ETTm2 and Weather (Supplementary Figs. 2-3), using progressively smaller training fractions (100 % down to 1 %) and evaluated the test MSE under each condition. To preserve the transferable temporal representations learned during pre-training, we froze all Transformer encoder parameters and updated only the channel-shared prediction head. This strategy keeps the pre-trained delay-subspace structure and attention dynamics intact while allowing the output layer to adapt efficiently to target-domain distributions with minimal supervision, thereby enhancing data efficiency and mitigating overfitting in low-resource or cross-domain settings.

As illustrated in Fig. 4, UDE-small demonstrates exceptional data efficiency. On the ETTm2 dataset, it surpasses state-of-the-art (SOTA) models trained directly on the full dataset (PatchTST[34] and iTransformer[35]) even without fine-tuning data, underscoring its robust pre-trained representations. On Weather, UDE-Small matches or exceeds SOTA performance with as little as 10 % of the training data and remains competitive when only 5 % is available (see also Supplementary Tables 1-4). These consistent gains across supervision levels highlight UDE's ability to encode general temporal dynamics via delay-subspace tokenization, enabling effective forecasting even under severe data limitations. Furthermore, we visualized the specific finetuned prediction results for ETTh1, ETTh2, Electricity and Traffic datasets using training data sets of 0%, 5%, 20% and 50% (see Supplementary Figs. 2-5).

Collectively, these results demonstrate UDE's remarkable fine-tuning efficiency and practical value for deployment in resource-constrained environments.

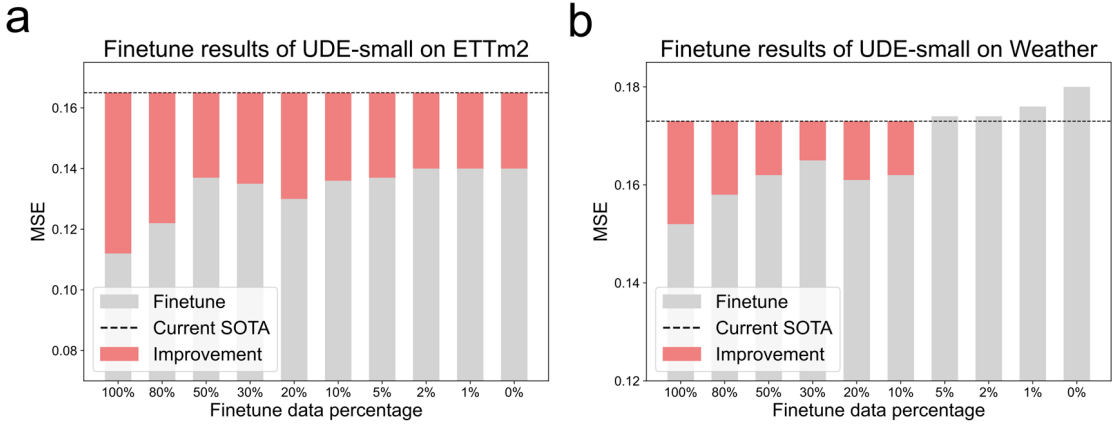

**Fig. 4. Data-efficiency of UDE-small on two benchmark datasets.** Fine-tuning results on **(a)** ETTm2 and **(b)** Weather are shown for labelled-data fractions ranging from 1 % to 100 %. The y-axis reports test MSE; the x-axis indicates the proportion of labelled data used for supervised fine-tuning. The black dashed line marks the best existing model trained directly on the target dataset without pre-training. Grey bars give UDE-Small's final performance after fine-tuning, and red segments denote their gain over SOTA. UDE-Small exceeds the baseline on ETTm2 even at 0 % fine-tuning and retains a Weather advantage down to 10 % training data.

**Representing nonlinear dynamics by UDE in the lifted latent space to approximate Koopman operator**

To elucidate the dynamical representations learned by UDE, we performed a series of visualization and scaling analyses, as depicted in Fig. 5. UDE functions as a finite-dimensional approximation of the Koopman operator (a linear operator designed to handle nonlinear dynamical systems) realized through delay-subspace tokenization combined with Transformer-based modelling. Here, we specifically explore whether this framework allows UDE to represent nonlinear chaotic or non-stationary dynamics in a lifted high-dimensional (latent) space in an approximately linear way.

In Fig. 5a, a pre-trained UDE encodes the Lorenz attractor (a canonical example of chaotic dynamics) into a latent space. Upon dimensionality reduction via PCA (green points), the encoded trajectories exhibit a nearly circular structure, effectively linearizing the attractor's characteristic "butterfly" shape into a quasi-periodic ring. Further visualization using t-SNE (orange points) preserves local neighborhood structures, clearly identifying three clusters corresponding to two metastable "wings" and their transitional regions. This underscores UDE's capability to disentangle complex dynamical modes purely from data.

To quantitatively assess how effectively UDE approximates the Koopman ideal, we varied its model width (hidden dimension) and evaluated the forecasting error after pre-training UDE-Small on the UTSD-1B[28] dataset. As shown in Fig. 5b, expanding the latent dimension from 16 to 128 monotonically reduces both mean squared error (MSE) and mean absolute error (MAE) toward theoretical minima across six benchmarks (ETTh1, ETTh2, ETTm1, ETTm2, Weather, Electricity). This indicates that larger models better capture Koopman-consistent linearization of latent state evolution.

Additionally, we assessed the structural properties of UDE's representation space. Fig. 5c illustrates t-SNE embeddings of samples from diverse domains (Electricity, Weather, Traffic), revealing well-separated clusters. This demonstrates UDE's ability to distinguish domain-specific dynamics without explicit supervision. Conversely, Fig. 5d shows substantial overlap among embeddings drawn from related datasets within a single domain (ETTh1, ETTh2, Electricity), highlighting UDE's extraction of domain-invariant features that facilitate effective intra-domain generalization.

These observations underscore a fundamental property of UDE as an approximate Koopman operator: its ability to represent nonlinear temporal dynamics within a latent space in which trajectories evolve approximately linearly. This capability stems from time-delay embedding theory, which reconstructs phase spaces via topological isomorphisms. By modelling tokens derived from TDP, UDE captures local geometric structures and facilitates global trajectory evolution through attention mechanisms. Thus, UDE learns an implicit mapping from the original nonlinear system to its Koopman representation, establishing linear or near-linear latent trajectories.

Importantly, this theoretical underpinning distinguishes UDE sharply from conventional sequence models such as RNNs, TCNs, and traditional Transformers, which operate directly on raw temporal or embedding spaces and rely on highly nonlinear mappings to extract temporal dependencies, typically lacking explicit topological guarantees or interpretable latent evolution. In contrast, UDE explicitly preserves phase-space geometry via delay embedding, aligning its representational strategy with a dynamical-systems perspective. This approach not only enhances forecasting accuracy but also provides interpretable, transferable, and structurally coherent representations.

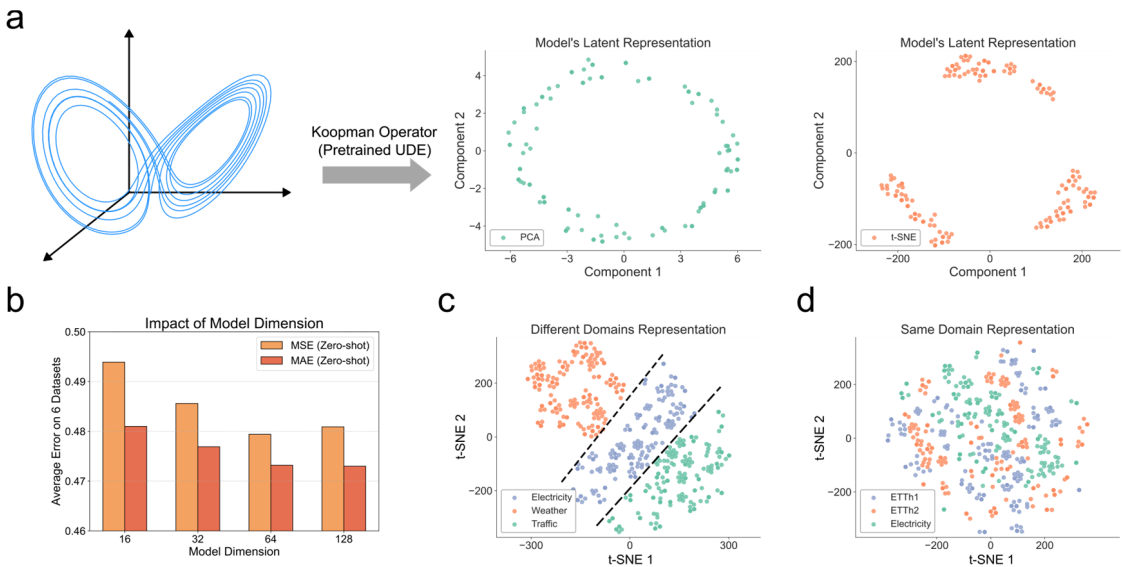

**Fig. 5 Representing nonlinear dynamics by UDE in the lifted latent space. (a)** A pre-trained UDE encodes the Lorenz attractor into a latent space that approximates a Koopman operator. PCA of the latent sequence (green dots) projects the nonlinear attractor onto an approximately circular ring, revealing its underlying quasi-periodic dynamics. t-SNE (orange dots) on the same embeddings yields three clearly separated clusters corresponding to the system's two metastable wings and the intervening transition region. **(b)** Expanding the latent dimension from 16 to 128 monotonically reduces both MSE and MAE in zero-shot forecasting across six datasets after training on UTSD-1B, indicating a progressively better Koopman approximation toward the theoretical optimum. **(c)** t-SNE visualizations of embeddings from distinct domains (Electricity, Weather, Traffic) form linearly separable clusters, demonstrating UDE's capacity to extract domain-specific

dynamics. **(d)** Conversely, embeddings from different datasets within the same domain (ETTh1, ETTh2, Electricity) exhibit substantial overlap, suggesting that UDE captures domain-invariant features when the underlying dynamics are structurally similar.

**UDE Performance on real-world city-level climate forecasting**

To further evaluate the practical applicability and generalization of UDE, we applied the pre-trained UDE-Small model to the ERA5[56] dataset, performing city-level climate forecasts for six representative urban areas: Beijing, Shanghai, Guangzhou, Tokyo, London, and Los Angeles. The forecast targets six meteorological variables: 2 m temperature, 2 m dew point temperature, 10 m U wind component, 10 m V wind component, mean sea level pressure, and surface pressure. We assessed UDE-Small under two scenarios: (i) zero-shot inference and (ii) fine-tuning on historical ERA5 data (1979–2018). Both settings were evaluated on recent ERA5 data (2019–2022) and compared against baseline models (PatchTST[34], iTransformer[35], DeepAR[57], Wavenet[58]) trained on the same historical period. The aggregated performance across all six variables is presented in Fig. 6a, whereas Fig. 6b–g illustrates representative zero-shot predictions of the 2 m temperature (t2m) variable. Detailed results are provided in the Supplementary Table 5-10.

In the challenging zero-shot setting, UDE-Small performs on par with or better than the baselines that were explicitly trained on ERA5 historical data, highlighting its strong domain-transfer capability. After fine-tuning, UDE-Small's performance improves markedly, outperforming all baselines and demonstrating its adaptability with limited domain-specific data. The t2m forecasts for 15–25 July 2023 (Fig. 6b–g) confirm that UDE effectively captures key meteorological patterns learned during pre-training, such as diurnal cycles and peak daily temperatures. Errors are concentrated during abrupt extreme-weather events or sudden temperature fluctuations, indicating clear directions for future refinement.

Taken together, these results establish that UDE's integration of time-delay embedding and latent Koopman-linearized representations enables robust forecasting in real-world nonlinear scenarios, offering distinct advantages in generalization.

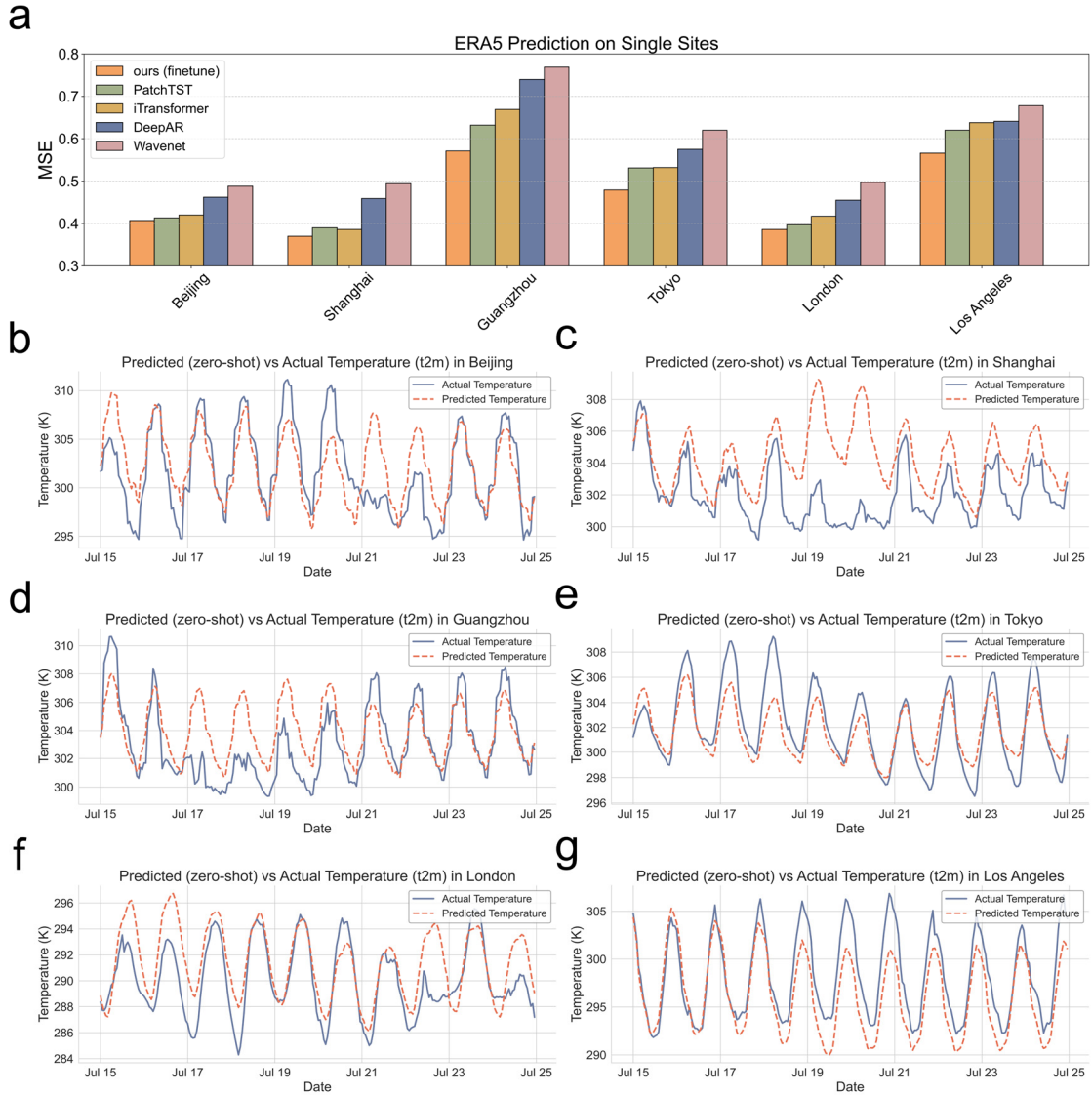

**Fig.6. City-level temperature forecasting with UDE-Small on the ERA5 dataset. (a)** Bar chart comparing MSE for 2019–2022 forecasts. UDE-Small, pre-trained on ERA5 data from 1979–2018, is evaluated in both zero-shot (light grey) and fine-tuned (dark grey) modes against established baselines (PatchTST, iTransformer, DeepAR, WaveNet) across six representative cities: Beijing, Shanghai, Guangzhou, Tokyo, London, and Los Angeles. UDE-Small achieves competitive or superior accuracy in zero-shot mode and improves markedly after fine-tuning. **(b–g)** Zero-shot dynamic forecasts (red dashed lines) versus observed 2 m temperature (blue solid lines) for 15–25 July 2023 in each city. UDE-Small consistently captures diurnal cycles and daily peaks; errors arise primarily during sporadic extreme-weather events.

## Discussion

This study introduces UDE, a pretrained foundational model with a delay-embedding architecture that models each time-series channel or variable as a dynamical representation of the observed high-dimensional data. Theoretically, UDE is grounded in rigorous dynamical-systems principles, thus endowing UDE with superior interpretability. Computationally, it partitions the corresponding Hankel matrix into two-dimensional subspace patches, which are tokenized and processed efficiently through self-attention mechanisms as images. Based on Takens' Embedding Theorem and STI equation, each patch functions as a localized projection of the reconstructed phase-space trajectory, enabling the encoder to implicitly approximate a finite-dimensional Koopman operator.

This architectural design offers four key advantages for time-series forecasting. UDE demonstrates exceptional zero-shot accuracy and remarkable data efficiency. Across seven standard benchmarks, it achieves the lowest average MSE, outperforming or matching significantly larger foundation models. Remarkably, using only 20% of available fine-tuning data, UDE attains SOTA results, surpassing previous best performances on the ETTm2 dataset even under zero-shot conditions. Second, the latent representations learned by UDE are inherently interpretable. For instance, a pretrained UDE successfully linearizes chaotic attractors (such as the Lorenz system) into near-circular manifolds. Visualization via t-SNE further reveals distinct dynamical modes. Moreover, attention mechanisms predominantly highlight a sparse subset of tokens that maintain topological coherence across Transformer layers and attention heads, providing stable reference points for identifying underlying dynamical structures. Third, UDE scales effectively as a Koopman operator approximation. Increasing the hidden layer width from 16 to 128 consistently reduces forecasting errors, indicating that wider models capture increasingly detailed Koopman spectral features. Finally, UDE exhibits robust forecasting capabilities in real-world scenarios. Applied to ERA5 city-level temperature data, UDE-small model performs comparably to established baselines under zero-shot conditions and significantly surpasses them after fine-tuning. It accurately captures essential meteorological patterns, including diurnal temperature cycles and extreme weather events.

The capability of UDE to linearize complex temporal dynamics stems fundamentally from its architectural foundation in delay embedding. According to Takens' theorem, delay embeddings reconstruct the underlying system's phase space as dynamical representations while preserving topological fidelity. Within UDE, this dynamical reconstruction occurs at the patch level, where each patch corresponds to a localized segment of the delay trajectory, preserving short-term dynamics and local geometric structures. By explicitly modeling transitions between these delay-subspace patches, UDE implicitly approximates a finite-dimensional Koopman operator. Consequently, the Transformer operates on latent representations with approximately linear dynamics. This structured approach diverges significantly from conventional black-box architectures, which typically must learn latent temporal dynamics directly from raw data. As Fig. 5d illustrates, increasing model width enhances UDE's spectral resolution, consistently improving forecasting accuracy. This Koopman-like characteristic explains UDE's robustness, data efficiency, and cross-domain generalization: Instead of initializing with arbitrary latent features, UDE begins from geometry-informed, physically consistent embeddings of the original dynamics.

Despite its performance advantages, UDE faces theoretical and practical constraints. First, its performance remains sensitive to the configuration of time-delay parameters, particularly the embedding dimension $m$ and patch partitioning scheme ($p$, $q$). Though $m = 500$ and $\tau = 1$ demonstrated robustness in pretraining, optimal values are empirically tuned rather than theoretically derived, potentially compromising attractor-reconstruction fidelity for signals with multiscale dynamics or nonstationary regimes. Second, computational overhead arises from Hankel matrix construction for long sequences. While token-pooling compression mitigates this burden, it risks discarding high-frequency dynamical nuances. Third, UDE's Koopman-linearization capability assumes smooth latent evolution, evidenced by its progressive error reduction with increased model width (Fig. 5b). This assumption fails for systems violating spectral compactness requirements, such as those exhibiting Lévy-flight characteristics or rapidly shifting attractors. These constraints delineate critical frontiers for adaptive parameterization and hybrid operator design.

While UDE demonstrates robust forecasting performance across diverse domains, we note that in scenarios with particularly high observational noise, predictions may be further stabilized through an optional aggregation scheme. In this approach, the prediction of a target variable is combined with outputs from other channels that exhibit strong dependencies with the target, quantified for example by mutual information. This consideration of cross-variable dependencies complements the system-level homomorphism ensured by delay embedding. Whereas Takens' theorem guarantees that a univariate delay embedding can reconstruct the global attractor geometry, the aggregation step explicitly leverages inter-variable relationships to suppress noise and enhance coherence across channels. Such an option provides a lightweight ensemble-like enhancement without modifying the UDE backbone and may be especially valuable when single-channel predictions are severely degraded by noise. To preserve our univariate per-channel paradigm, all results in the main text do not use cross-channel fusion. For completeness, we additionally report an optional test-time, target-only mutual information-guided late aggregation that blends the target forecast with its Top-5 neighbors using fixed weights after z-score alignment (Supplementary Note 5).

Future research could extend UDE's theoretical depth and practical scope through several promising avenues. First, topological persistence analysis could dynamically tune hyperparameters using Wasserstein distances between learned token manifolds, enhancing reconstruction fidelity for nonstationary systems[59]. Second, sparse factorization of Hankel matrices using randomized numerical methods could reduce memory complexity while maintaining geometric integrity via topological regularization[60]. Third, integrating UDE's delay-aware tokens with vision-language models (VLMs)[61,62] might enable unified spatiotemporal forecasting, such as generating typhoon trajectories conditioned on satellite imagery and textual weather reports. Finally, addressing real-world deployment challenges highlighted in climate applications will likely require hybrid architectures. For instance, coupling UDE with anomaly or tipping-point detection modules could enhance robustness during extreme events, advancing toward physics-informed foundation models.

## Method

**Problem formulation**

Let $x^t = [x_1^t, ..., x_n^t]' \in \mathbb{R}^n$ be an $n$-channel time series observed for $t = 1, 2, ..., T$. Given a forecasting horizon $h$, we aim to make $h$-step prediction

$$x^{T+1}, ..., x^{T+h}$$

under two scenarios: (i) zero-shot: the model is used exactly as pretrained and no target data updates, and (ii) fine-tune: a small, task-specific split of the target data is used for supervised adaptation.

**Time-delay embedding and Hankel matrix as a dynamical representation**

For each variable $x_i^t$ we build a time-delay embedding vector or Eq. (1): $y_i^t = \left[x_i^{t-(m-1)\tau}, ..., x_i^{t-2\tau}, x_i^{t-\tau}, x_i^t\right]' \in \mathbb{R}^m$ with delay step $\tau = 1$ and embedding dimension $m$. Based on the delay embedding theorem or STI equation with a few generic conditions, there is generally a smooth function $\phi$ satisfying $y_i^t = \phi_i(x^t)$ or $x^t = \phi_i^{-1}(y_i^t)$, which is clearly a universal representation form or a universal approximation framework, i.e. by a sequence $y_i$ representing a vector $x$. Actually, such a delay-embedded representation $y_i^t$ can be viewed as a dynamical representation[39–41] of the original $x^t$ in contrast to the static representation used widely in deep learning. By a nonlinear map (e.g. a neural network), $y_i^t$ can be further transformed to $z_i^t = [z_i^t, z_i^{t+1}, \dots, z_i^{t+M-1}]' \in \mathbb{R}^M$. Stacking all valid delay vectors yields a Hankel matrix

$$H_i = \begin{bmatrix} y_i^{m'} \\ y_i^{m+1'} \\ \vdots \\ y_i^{m+L-1'} \end{bmatrix} = \begin{bmatrix} x_i^1 & x_i^2 & \cdots & x_i^m \\ x_i^2 & x_i^3 & \cdots & x_i^{m+1} \\ \vdots & \vdots & \ddots & \vdots \\ x_i^L & x_i^{L+1} & \cdots & x_i^{m+L-1} \end{bmatrix}, L = T - m + 1$$

whose rows reconstruct the system's local phase–space trajectory while retaining the original topology (Takens' theorem). A moderate value of $m$ balances expressiveness and computational tractability. In the pretraining process, we empirically set $m = 500$ to make it above the dominant temporal cycles on most datasets. $H_i$ can be viewed as a form of dynamical representation $y_i$ for state $x$, which transforms a state vector $x$ at time $t$ to a time series or sequence at time $t, t + 1, \dots$.

**Time-delay patching**

Each $H_i$ is partitioned into non-overlapping 2D patches of size $p \times q$. $P_i^{(j)} = H_i[(u-1)p + 1: up, (v-1)q + 1: vq]$, where $u = 1, \dots, U$, $U = \lfloor L/p \rfloor, v = 1, \dots, V, V = \lfloor m/q \rfloor$ and $j = (u-1)V + v$. Thus, there are $M = U \times V$ patches in total. Each patch spans $p$ consecutive time steps and $q$ delayed coordinates, forming a localized subspace projection of the full delay embedding that preserves both temporal and geometric structure.

Here is a more specific example. Let a time series $x_i^t$ be observed for $T = 9$ time steps with an embedding dimension $m = 4$. The resulting Hankel matrix is:

$$H_i = \begin{bmatrix} x_i^1 & x_i^2 & x_i^3 & x_i^4 \\ x_i^2 & x_i^3 & x_i^4 & x_i^5 \\ x_i^3 & x_i^4 & x_i^5 & x_i^6 \\ x_i^4 & x_i^5 & x_i^6 & x_i^7 \\ x_i^5 & x_i^6 & x_i^7 & x_i^8 \\ x_i^6 & x_i^7 & x_i^8 & x_i^9 \end{bmatrix} \in \mathbb{R}^{6 \times 4}$$

Now, setting patch size $p = 2$, $q = 2$, the matrix is divided into $U = 3$ and $V = 2$, thus $M = 6$ patches in total. For example: $P_i^{(1)} = \begin{bmatrix} x_i^1 & x_i^2 \\ x_i^2 & x_i^3 \end{bmatrix}$, $P_i^{(2)} = \begin{bmatrix} x_i^3 & x_i^4 \\ x_i^4 & x_i^5 \end{bmatrix}$, each patch is thus a two-dimensional local view capturing time-delay dynamics in a compact form. The patch size controls the temporal and geometric resolution of the learned representations. In our design $p$ governs the number of time steps captured per patch, while $q$ determines the delay span. Practically, we select values $p = 25$ and $q = 50$ to ensure that patches capture local dynamic behavior while allowing the full Hankel matrix to be covered with minimal information loss.

**UDE Architecture**

Each time-delay patch $P_i^{(j)}$ is first flattened into a vector

$$\mathbf{P}_i^{(j)} = \text{vec}\left(\mathbf{P}_i^{(j)}\right) \in \mathbb{R}^{pq}.$$

Then linearly token embedding projected into a latent space

$$\mathbf{e}_i^{(j)} = \mathbf{W}_e \mathbf{P}_i^{(j)} + \mathbf{b}_e \in \mathbb{R}^d,$$

where $\mathbf{W}_e \in \mathbb{R}^{d \times pq}$ is a learnable projection matrix. The full patch sequence forms:

$$\left[\mathbf{e}_i^{(1)}, \dots, \mathbf{e}_i^{(M)}\right]' \in \mathbb{R}^{M \times d}.$$

Since it has been proven that the tokens adjacent tokens have properties, similar to images, token pooling for down-sampling was used to reduce computational cost while retaining coarse global structure:

$$\mathbf{E} = \text{AvgPool1D}\left(\left[\mathbf{e}_i^{(1)}, \dots, \mathbf{e}_i^{(M)}\right]', \text{kernel} = k, \text{stride} = s\right),$$

where $k = s = 30$ in our default configuration. Sinusoidal positional encoding $\mathbf{PE} \in \mathbb{R}^{M \times d}$ is added:

$$\mathbf{X}^{(0)} = \mathbf{E} + \mathbf{PE}.$$

UDE utilizes an encoder-only transformer structure, and a sinusoidal positional encoding instead of learnable positional encoding for stability. The projected tokens are processed through a Transformer encoder. Each block is composed of multi-head self-attention and MLP sublayers with residual connections. Let $\mathbf{X}^{(l)}$ be the input at layer $l$. Each encoder layer includes overs two steps, one multi-head self-attention (MHA) layer and one feed-forward (FF) layer:

$$\mathbf{A}^{(l)} = \text{LayerNorm}\left(\mathbf{X}^{(l)} + \text{MHA}(\mathbf{X}^{(l)})\right)$$
$$\mathbf{X}^{(l+1)} = \text{LayerNorm}\left(\mathbf{A}^{(l)} + \text{FF}(\mathbf{A}^{(l)})\right).$$

After the final encoder layer, the entire sequence $\mathbf{X}^{(n)}$ was flattened into a single vector and then passed through a prediction head:

$$\hat{X} = \text{flatten}(\mathbf{X}^{(n)})\mathbf{W} + \mathbf{b}.$$

Three configurations are used for evaluation: UDE-Small (6 layers, 8 heads); UDE-Medium (8 layers 12 heads); UDE-Large (12 layers 16 heads). To enhance computational efficiency, a 1-D average pooling layer with kernel and stride set to 30 is applied before attention layers. This reduces the number of input tokens while preserving coarse global structure. A one-layer MLP head with GELU activation is used to decode the sequence representation. The forecast head maps latent representations to the predicted $H$-step future outputs.

**Pretraining**

For large-scale pretraining, we use the USTD-12B dataset (12 billion steps across 10 domains), while USTD-1B is used for ablation and validation studies. The model is trained to minimize MSE using the Adam optimizer with a learning rate of 0.001 and batch size of 16,384. Cosine learning rate annealing is applied every epoch. All pretraining is performed on Nvidia Tesla A800 (80G) GPUs. Average pooling with kernel and stride set to 30 is applied for token reduction prior to Transformer encoding.

**Dynamical Representation and Static Representation**

Conventional deep learning models for time-series forecasting typically employ a static representation of the input sequence (Fig.2f). In this setting, the feature mapping $\Phi$ is applied

independently to the observation vector $x^t \in \mathbb{R}^n$ at each time step:
$$z^t = \Phi(x^t) \text{ or } x^t = \Phi^{-1}(z^t), t = 1,2,...$$
where $z^t \in \mathbb{R}^k$ denotes the latent feature vector $[z_1^t, z_2^t, ..., z_k^t]$ at time $t$. The static representation implies that a latent vector (e.g. $z^t = [z_1^t, z_2^t, ..., z_k^t]'$) with $k$ variables corresponds to a state or vector (e.g. $x^t = [x_1^t, x_2^t, ..., x_n^t]'$) of the observed different $n$ variables but all at the same time point $t$ (see Fig. 2f). This representation treats each time step as an isolated sample and does not explicitly encode temporal continuity or local geometry of the underlying trajectory in state space. Consequently, temporal dependencies must be learned implicitly by subsequent sequence models usually not in a dynamical but in a statistic manner (e.g., RNNs, Transformers).

In contrast, dynamical representation directly incorporates local temporal context into the encoding stage. Specifically, we construct a time-delay embedding of length $m$ or $M$ with a map $\Phi$ from consecutive observations:
$$z_i^t = \Phi(x^t) \text{ or } x^t = \Phi^{-1}(z_i^t), t = 1,2,...$$
Here $z_i^t \in \mathbb{R}^M$ (or $y_i^t$ in Eq. (1) denotes a scalar sequence as $[z_i^t, z_i^{t+1}, ..., z_i^{t+M-1}]$. Dynamical representation means that a sequence or latent vector $z_i^t = [z_i^t, z_i^{t+1}, ..., z_i^{t+M-1}]'$ or $y_i^t$ of an observed single variable $x_i$ but with different time points $(t - (m-1), ..., t+1, t)$ represents a state or vector $x^t = [x_1^t, x_2^t, ..., x_n^t]'$ of the observed different $n$ variables at the same time point $t$ (see Fig. 2e). Note that we use $M$ instead of the dimension $m$ in Eqs. (4)-(6) for $z$ due to the dimension lifting from $y$ to $z$.

By Takens' embedding theorem with a few generic conditions, we can prove that such delay-embedded segments are topologically equivalent to the original system's attractor[41-45], thereby preserving both the geometric structure and the dynamical properties of the underlying process. Therefore, it can also be viewed as a universal representation form of the observed high-dimensional data from the viewpoint of dynamical systems.

**Topological analysis**

To quantify structural similarity between time-delay patches, we adopt persistent homology from topological data analysis (TDA). Each patch is treated as a point cloud in delay embedding space, where temporal sequences are transformed into high-dimensional vectors through $y^t$. Following Takens' embedding theorem, this reconstruction preserves the system's dynamical structure in geometric form. To extract topological signatures, we constructed Vietoris–Rips simplicial complexes across a sequence of distance thresholds $\epsilon$, forming a filtration:
$$\emptyset = K_0 \subseteq K_1 \subseteq \cdots \subseteq K_M$$
At each level, we identify homological features: $H_0$ (connected components) and $H_1$ (loops). Their appearance and disappearance across the filtration are encoded in persistence diagrams, where each point $(b, d)$ denotes a topological feature born at scale $b$ and dying at $d$.

To compare patch-wise topological structure, we compute the 2-Wasserstein distance between their persistence diagrams:
$$W_p(D_1, D_2) = \left( \inf_{\gamma \in \Gamma} \sum_{(u,v) \in \gamma} \| u - v \|_\infty^p \right)^{1/p}$$
Here, $\Gamma$ is the set of partial matchings (including diagonal projections), and $\|\cdot\|_\infty$ is the max norm. This yields a geometry-informed distance metric that captures intrinsic similarities across delay-embedded subspaces. Notably, block structures observed in the Wasserstein distance matrix (e.g., Fig. 3c) reflect latent topological coherence among token groups, supporting the hypothesis that UDE patches reside on a shared manifold shaped by underlying dynamical modes.

## Supplementary information

Supplementary information (Supplementary Notes 1~5, Supplementary Tables 1~10 and Supplementary Figs. 1~5) is available for this paper.

## Data and Code availability

The UDE algorithm was implemented in Python and is available on GitHub at https://github.com/Wangzj000/UDE. All data utilized in this study derive from publicly accessible sources.

## Acknowledgments

We thank all the members of Chen laboratory for technical assistance. This work has been supported by National Key R&D Program of China (2022YFA1004800, 2025YFF1207900), Natural Science Foundation of China (T2341007, T2350003, 12131020, 12301620, 42450084, 42450135, 42450192, 12326614, and 12426310), Science and Technology Commission of Shanghai Municipality (23JS1401300), Zhejiang Province Vanguard Goose-Leading Initiative (2025C01114), Hangzhou Institute for advanced study of UCAS (2024HIAS-P004), JST Moonshot R&D (JPMJMS2021), the Shanghai Municipal Data Bureau special fund for urban digital transformation (202401065), and the AI for Science Foundation of Fudan University (FudanX24AI041).

## Author contributions

Z.J.W, P.T. and L.N.C. conceived the idea; Z.J.W. and P.T. designed the research; Z.J.W. and P.T. performed the research. All authors analyzed the data and wrote the paper.

## Competing interests

The authors declare that they have no competing interests.

# Supplementary Information

for

# A Time-Series Foundation Model by Universal Delay Embedding

# Supplementary Note 1. Implementation Details

## 1.1 Datasets used in the main text

We conducted experiments on seven widely used real-world multivariate time-series forecasting datasets, as well as one high-resolution real-time climate dataset, ERA5, as described in the main text. **ETT** (Electricity Transformer Temperature) datasets are collected from electricity transformers at two locations and labeled with 1 and 2, covering 7 variables. ETT datasets contain 4 datasets: **ETTh1**, **ETTh2**, **ETTm1**, and **ETTm2**, where h and m represent two sampling intervals: 1 hour (h) and 15 minutes (m). **Weather** dataset collects 21 meteorological indicators every 10 minutes from Germany (e.g., humidity, air temperature) in 2020. **Electricity** dataset records the electricity consumption every 1 hour from 321 customers. **Traffic** dataset contains road occupancy rates from 862 sensors in San Francisco.

ERA5 is a state-of-the-art global climate reanalysis dataset developed by the European Centre for Medium-Range Weather Forecasts (ECMWF) under the Copernicus Climate Change Service (C3S). It offers hourly estimates of numerous atmospheric, land, and oceanic variables from 1950 to the present, with continuous updates. For our study, we selected hourly data on single levels from the ERA5 archive, covering six key variables: 2 m temperature, 2 m dewpoint temperature, 10 m U-component wind, 10 m V-component wind, mean sea level pressure, and surface pressure.

## 1.2 Workflow of UDE

UDE framework introduces a structured pipeline that integrates dynamical systems theory with Transformer-based deep learning for universal time-series forecasting. The overall workflow consists of the following major stages:

### 1.2.1 Input and Delay Embedding Construction

Given a multivariate time series $X \in \mathbb{R}^{T \times C}$ (typically a sliding window), where $T$ is the temporal length and $C$ is the number of channels, the goal is to forecast a horizon $h$, i.e., predict $\hat{Y} \in \mathbb{R}^{h \times C}$, in either zero-shot or fine-tuning settings.

For each channel, a delay embedding is constructed using fixed parameters $m$ (embedding dimension) and $\tau$ (delay step, typically $\tau = 1$):

$$\boldsymbol{y}_i(t) = [x_i(t), x_i(t+\tau), x_i(t+2\tau), \ldots, x_i(t+(m-1)\tau)] \in \mathbb{R}^m.$$

These embeddings are stacked to form a Hankel matrix $\boldsymbol{H} \in \mathbb{R}^{(T-m+1) \times m}$.

### 1.2.2 Time-Delay Patching

The Hankel matrix is partitioned into non-overlapping rectangular 2D patches of size $p \times q$, resulting in $\lfloor (T-m+1)/p \rfloor \times \lfloor m/q \rfloor$ total patches. Each patch encodes a localized projection of the system's trajectory in the delay-embedded space, maintaining both temporal and geometric consistency.

### 1.2.3 Patch Projection and Transformer Encoding

Each 2D patch is flattened and projected to a latent space of dimension $d$ via a linear layer. A sinusoidal positional encoding is added (fixed, non-learnable), and the resulting sequence of tokens is passed through a Transformer encoder consisting of multiple layers of multi-head self-attention and MLP blocks. A token pooling layer (1D average pooling, kernel = stride = 30) is applied to reduce sequence length and promote global structure awareness.

### 1.2.4 Forecast Head and Output

After Transformer encoding, the resulting patch-level token representations are aggregated and passed through a shared, channel-invariant forecast head. Specifically, all channels share the same MLP decoder, which maps latent tokens to a forecast vector of length $h$ (prediction horizon). This design enforces cross-channel parameter sharing, improving parameter efficiency and promoting inductive bias across heterogeneous variables. The final output is a multivariate forecast $\hat{Y}$, reconstructed by repeating the shared MLP across all channels.

### 1.3 Training details

As described in the main text, we constructed three UDE model variants of different sizes. The corresponding architectural configurations and hyperparameters are summarized in the table below. Following the findings of Delayformer, which demonstrated that the choice of positional encoding has minimal impact on forecasting performance, we adopted fixed sinusoidal positional encoding for simplicity, although learnable positional embeddings are commonly used in Vision Transformer architectures.

Across all experiments, the time-delay embedding dimension and patching configuration were fixed at $m = 500$, $p = 25$, and $q = 50$. Layer Normalization and a dropout rate of 0.1 were consistently applied during both pretraining and fine-tuning. Model training was performed using the Adam1 optimizer with cosine learning rate annealing. The learning rate was set to $1 \times 10^{-4}$ during pretraining and reduced to $5 \times 10^{-5}$ for fine-tuning. The batch size was set to 16384. During the fine-tuning process, we utilized varying percentages of the training data, sampled uniformly from the available dataset. All the experiments used in the main text were implemented with PyTorch on a four NVIDIA A800 80GB GPUs.

|  | Layers | Heads | $d_{model}$ | $d_{ff}$ | Total Params |
|---|---|---|---|---|---|
| UDE-small | 6 | 8 | 512 | 2048 |  |
| UDE-medium | 10 | 12 | 512 | 2048 |  |
| UDE-large | 12 | 16 | 512 | 2048 |  |

# Supplementary Note 2. Topological Data Analysis for Delay-Embedded Time Series

To characterize the geometric and topological structure of time-delay embedded sequences, we applied Topological Data Analysis (TDA) based on persistent homology and Wasserstein distance. This approach enables us to quantify the intrinsic dynamic features of multivariate time series beyond raw signal similarity. For implementation, we used GUDHI[2] library.

## 2.1 Delay Embedding and Point Cloud Construction

Let $x(t)$ be a univariate time series. Following Takens' embedding theorem, we reconstruct the system's phase space using delay embedding vectors:

$$\mathbf{y}(t) = [x(t), x(t-\tau), x(t-2\tau), \ldots, x(t-(d_E-1)\tau)]$$

where $\tau$ is the time delay, and $d_E$ is the embedding dimension. Each sequence is thus transformed into a point cloud in $\mathbb{R}^{d_E}$, preserving the underlying dynamics in geometric form[3].

## 2.2 Persistent Homology and Simplicial Complexes

To extract topological features, we construct a Vietoris–Rips complex over the point cloud for a sequence of distance thresholds $\epsilon$, forming a filtration:

$$\emptyset = K_0 \subseteq K_1 \subseteq \cdots \subseteq K_M$$

At each scale, simplices are formed among points whose pairwise distances are within $\epsilon$. This yields topological invariants, homology groups, $H_k$ capturing features such as:

$H_0$ (connected components); $H_1$ (loops or cycles); $H_2$ (voids)[4].

We track the birth and death of these features across the filtration to form a persistence diagram, where each point $(b, d)$ corresponds to a feature born at scale $b$ and disappearing at $d$. Features with high persistence $(d - b)$ reflect significant structure; those near the diagonal are considered noise.

## 2.3 Wasserstein Distance Between Persistence Diagrams

To compare the topological structure of two time series, we compute the Wasserstein distance $W_p$ (typically $p = 2$) between their persistence diagrams[5]:

$$W_p(D_1, D_2) = \left( \inf_{\gamma \in \Gamma} \sum_{(u,v) \in \gamma} \| u - v \|_\infty^p \right)^{1/p}$$

Here, $\gamma$ is a matching between points in diagrams $D_1$ and $D_2$, allowing unmatched points to be mapped to the diagonal. $\|\cdot\|_\infty$ is the maximum norm.

## 2.4 Total Weighted Wasserstein Distance (TWWD)

To obtain a single summary distance across dimensions, we compute the total weighted Wasserstein distance[6]:

$$\text{TWWD}(X_1, X_2) = \left( \sum_{k=0}^{k_{max}} w_k \cdot W_p(D_1^{(k)}, D_2^{(k)})^p \right)^{1/p}$$

where $w_k$ is a normalized weight derived from the mean persistence of finite features in dimension $k$. This ensures that more structurally informative dimensions contribute proportionally to the final distance.

## 2.5 Application and Interpretation

Applied to the delay-embedded patches in our Universal Delay Embedding (UDE) framework, this TDA pipeline enables the comparison of subspace structures. For instance, in Fig. 3c of the main text, we show Wasserstein distances between all token pairs derived from a delay embedding. Clear block structures in the resulting matrix reveal groups of tokens with similar topological patterns, corresponding to distinct dynamical modes. These findings support the interpretability of UDE token space and validate the use of geometry-aware patching in modeling time series dynamics.

## Supplementary Note 3. Subspaces in UDE

Denote the column vectors of patch $P_i^{(j)}$ as $\{v_1, ..., v_q\} \in \mathbb{R}^p$, then the column space $\mathcal{S}_{ij} = \text{span}\{v_1, ..., v_q\} \in \mathbb{R}^p$ defines a local low-dimensional linear subspace. If $q \ll p$, this subspace approximates the directionality of local dynamics. The similarity between subspaces can be measured by canonical angles or projection metrics. These local subspaces can be interpreted as projections of the global manifold $\mathcal{M}$, and their topology reflects properties such as periodicity, chaos, or drift.

For structural analysis, each patch can be interpreted as a point cloud in $\mathbb{R}^p$ (along columns), and topological invariants (e.g., 1D loops) can be extracted using persistent homology. Subspace similarity can be quantified by 2-Wasserstein distance between their persistence diagrams:

$$W_2(P_{ij}, P_{kl}) = \text{Wasserstein}_2(\text{Dgm}(P_{ij}), \text{Dgm}(P_{kl}))$$

This quantifies how similar two local dynamical regimes are in terms of topological structure.

Each patch $P_i^{(j)} \in \mathbb{R}^{p \times q}$ is flattened to a vector $z_i^{(j)} \in \mathbb{R}^{pq}$, then projected into latent space via a learnable linear mapping:

$$\mathbf{e}_{ij} = W(vec(P_i^{(j)})) + b \in \mathbb{R}^d$$

This latent embedding $\mathbf{e}_{ij}$ serves as a token in the Transformer encoder, encoding the subspace's dynamical structure and geometric properties.

# Supplementary Note 4. Delay Embedding Theory in Dynamical System

Delay embedding is a foundational technique for reconstructing the state space of a dynamical system from a univariate time series. This approach enables the recovery of the underlying attractor geometry using only time-delayed observations of a single scalar variable. The theoretical foundation rests on the Takens' Embedding Theorem, which guarantees the preservation of topological and dynamical invariants under specific conditions.

## 4.1 Dynamical System and Observations

Let $\mathcal{M} \in \mathbb{R}^n$ be a compact $d$-dimensional smooth manifold representing the state space of a deterministic dynamical system governed by a smooth diffeomorphism $D: \mathcal{M} \to \mathcal{M}$. Let $\phi_t$ be the flow generated by $D$, i.e., $\boldsymbol{x}^{t+1} = D(\boldsymbol{x}^t)$.

Only a scalar observation function $h: \mathcal{M} \to \mathbb{R}$ is available, and the observable time series is given by:

$$x^t = h(\boldsymbol{x}^t), t = 0,1,2,\ldots$$

## 4.2 Delay Coordinate Map

Given a scalar time series $x^t$, construct the delay embedding using $m$ delayed samples with lag $\tau \in \mathbb{N}$:

$$\Psi_{h,\tau,m}(x) = \left[h(x), h(D^{-\tau}(x)), \ldots, h\left(D^{-(m-1)\tau}(x)\right)\right]' \in \mathbb{R}^m$$

Each embedded point $\boldsymbol{y}^t \in \mathbb{R}^m$ corresponds to:

$$\boldsymbol{y}^t = \left[x^t, x^{t-\tau}, \ldots, x^{t-(m-1)\tau}\right]$$

This defines a map $\Psi: \mathcal{M} \to \mathbb{R}^m$ from the original manifold to the reconstructed delay space.

## 4.3 Takens' Embedding Theorem

Takens' theorem provides sufficient conditions for the reconstruction of a diffeomorphic image of the attractor from scalar observations:

Let $\mathcal{M}$ be a compact smooth manifold of dimension $d$, and let $D$ be a generic smooth diffeomorphism on $\mathcal{M}$. Then, for a generic smooth observation function $h: \mathcal{M} \to \mathbb{R}$, the map

$$\Psi_{h,\tau,m}: \mathcal{M} \to \mathbb{R}^m$$

is an embedding (i.e., a smooth diffeomorphism onto its image) provided that:

$$m \geq 2d + 1$$

The map $\Psi$ preserves the topological invariants of the original attractor $\mathcal{A} \subset \mathcal{M}$, including dimension (e.g., fractal, box-counting), connectedness and loop structures, qualitative trajectories (e.g., periodic orbits, chaotic sets). This means the reconstructed set $\Psi(\mathcal{A}) \subset \mathbb{R}^m$ is homeomorphic to the original attractor, and hence suitable for downstream tasks such as forecasting, classification, or topological analysis.

In real applications, $\tau$ can be selected via autocorrelation or mutual information criteria to avoid redundant sampling. $m$ is chosen heuristically or via methods such as false nearest neighbors (FNN). In UDE, we fix $\tau = 1$ and select large enough $m$ (e.g., $m = 500$) to empirically guarantee embedding in high dimensions.

## 4.4 Koopman Operator as a Linearization Framework for Nonlinear Dynamics Prediction

While delay embedding reconstructs the attractor geometry, the Koopman operator provides a linear framework for representing nonlinear dynamics.

For the dynamical system $x^{t+1} = D(x^t)$, the Koopman operator $\mathcal{K}$ acts on observables $g: M \to \mathbb{R}$ as

$$(\mathcal{K}g)(x) = g(D(x)).$$

Although $D$ may be nonlinear, $\mathcal{K}$ is linear on the space of observables. This linearity enables the modeling of nonlinear system evolution in a higher-dimensional (possibly infinite-dimensional) functional space.

In the UDE framework, the delay-embedded vectors $y^t$ are lifted to a high-dimensional latent space via a nonlinear encoder (e.g., Transformer) $\mathbb{R}^m \to \mathbb{R}^M$. Within this lifted space, temporal evolution is approximated by a **linear map** $K$, or $\Phi$:

$$z^{t+1} \approx \Phi(z^t)$$

Where $z^t$ denotes the latent representation. For an $h$-step prediction,

$$z^{t+1} \approx \Phi^h z^t.$$

Given latent trajectories $\{z^t\}$, the least-squares estimate minimizing $\sum_t \| z^{t+1} - Kz_t \|_2^2$ is $K^\star = Z_+ Z_-^\dagger$, $Z_+ = [z^1, \ldots, z^T]$, $Z_- = [z^0, \ldots, z^{T-1}]$. In UDE, $K$ is learned end-to-end together with $\Phi$, aligning the latent features with a Koopman-consistent subspace. If $K = V\Lambda V^{-1}$ (diagonalizable), then $z^{t+h} = V\Lambda^h V^{-1} z^t$, stability and periodicity corresponding to $|\lambda_i|$ and $\arg(\lambda_i)$. Increasing $M$ enlarges the span available to approximate Koopman eigenfunctions, improving linearization quality (as observed in UDE scaling experiments).

# Supplementary Note 5. Mutual Information Guided Late Aggregation

As discussed in the main text, predicting a target variable under noisy conditions can be stabilized by leveraging other variables that exhibit strong statistical dependence with the target, quantified by mutual information (MI). We implement this as a target-only enhancer that leaves the UDE backbone unchanged and maintains channel-invariant design (shared prediction head), consistent with the framework and discussion in the main text. These forecasts for a small set of auxiliary channels are computed internally at test time solely to assist the target via late aggregation; they are not reported as outputs and do not affect training. This preserves the univariate forecasting paradigm for per channel in the main text results.

Given a dataset with channel $1, 2, \ldots, N$ and a chosen target $k$, we compute pairwise MI $I(x^{(i)}; x^{(k)})$ for $i \neq k$ between the target and each other channel on the training split only and MI is calculated as follows. We then rank auxiliaries by (normalized) MI and keep the Top-5: $\mathcal{J}_k = \text{Top5}\{\text{NMI}(x_i; x_k) : i \neq k\}$.

Let $f_{XY}(x, y)$ be the joint density of $(X, Y)$ with marginals $f_X(x)$ and $f_Y(y)$. The (differential) entropies are

$$h(X) = -\int f_X(x) \log f_X(x) dx, \; h(Y) = -\int f_Y(y) \log f_Y(y) dy, \; h(X, Y)$$
$$= -\int \int f_{XY}(x, y) \log f_{XY}(x, y) dx dy,$$

and the mutual information (MI) is

$$I(X; Y) = \int \int f_{XY}(x, y) \log \frac{f_{XY}(x, y)}{f_X(x) f_Y(y)} dx dy = h(X) + h(Y) - h(X, Y)$$
$$= D_{\text{KL}}(f_{XY} \| f_X f_Y) \geq 0,$$

with natural logarithms.

Since differential entropies can be unbounded, we form a bounded normalized MI by computing entropies on a discretized (histogram) version of the z-scored training data:

$$\text{NMI}(X;Y) = \frac{I(X;Y)}{\sqrt{H_b(X)H_b(Y)}} \in [0,1]$$

where $H_b$ is Shannon entropy, continuous version of discrete entropy ($h$).

Let $\hat{y}_k^{t+h}$ be the baseline forecast of the target at horizon $h$, and $\hat{y}_i^{t+h}$ the auxiliary forecasts for the selected neighbors. The final target-only prediction is the fixed-weight blend with $|\mathcal{J}_k| = 5$ so that weights sum to 1:

$$\tilde{y}_k^{t+h} = 0.9\hat{y}_k^{t+h} + 0.02 \sum_{j \in \mathcal{J}_k} \mathcal{A}_{kj}(\hat{y}_i^{t+h}).$$

$\mathcal{A}_{kj}$ is a z-score alignment operator, which introduces no learned calibrator, uses training-split statistics only (no leakage), and works for both homogeneous-unit and heterogeneous-unit datasets.

All main-text tables are reported without this enhancer. In the Weather dataset, we applied the enhancement to the OT (target) variable and observed the prediction MSE improve from 0.101 to 0.098 (~3% relative). Similar target-only enhancements can be applied to other datasets following the same protocol.

## Supplementary Tables

**Supplementary Table 1.** Full results (MSE) of fine-tuning on ETTm2 dataset

| Data Percentage | UDE-small | UDE-medium | UDE-large |
| --- | --- | --- | --- |
| 100% (Full-shot) | 0.112 | 0.113 | 0.114 |
| 80% | 0.122 | 0.122 | 0.126 |
| 50% | 0.137 | 0.138 | 0.138 |
| 30% | 0.135 | 0.139 | 0.138 |
| 20% | 0.130 | 0.131 | 0.134 |
| 10% | 0.136 | 0.138 | 0.141 |
| 5% | 0.137 | 0.139 | 0.141 |
| 2% | 0.140 | 0.142 | 0.144 |
| 1% | 0.140 | 0.141 | 0.143 |
| 0% (Zero-shot) | 0.140 | 0.142 | 0.144 |

**Supplementary Table 2.** Full results (MAE) of fine-tuning on ETTm2 dataset

| Data Percentage | UDE-small | UDE-medium | UDE-large |
| --- | --- | --- | --- |
| 100% (Full-shot) | 0.230 | 0.232 | 0.232 |
| 80% | 0.240 | 0.242 | 0.244 |
| 50% | 0.257 | 0.258 | 0.258 |
| 30% | 0.255 | 0.260 | 0.259 |
| 20% | 0.252 | 0.253 | 0.254 |
| 10% | 0.257 | 0.259 | 0.262 |
| 5% | 0.258 | 0.259 | 0.262 |
| 2% | 0.262 | 0.263 | 0.265 |
| 1% | 0.262 | 0.262 | 0.265 |
| 0% (Zero-shot) | 0.262 | 0.263 | 0.265 |

**Supplementary Table 3.** Full results (MSE) of fine-tuning on Weather dataset

| Data Percentage | UDE-small | UDE-medium | UDE-large |
|---|---|---|---|
| 100% (Full-shot) | 0.152 | 0.154 | 0.153 |
| 80% | 0.158 | 0.159 | 0.159 |
| 50% | 0.162 | 0.162 | 0.162 |
| 30% | 0.165 | 0.163 | 0.164 |
| 20% | 0.161 | 0.161 | 0.163 |
| 10% | 0.162 | 0.161 | 0.163 |
| 5% | 0.174 | 0.175 | 0.176 |
| 2% | 0.174 | 0.177 | 0.180 |
| 1% | 0.176 | 0.177 | 0.181 |
| 0% (Zero-shot) | 0.180 | 0.184 | 0.187 |

**Supplementary Table 4.** Full results (MSE) of fine-tuning on Weather dataset

| Data Percentage | UDE-small | UDE-medium | UDE-large |
|---|---|---|---|
| 100% (Full-shot) | 0.209 | 0.210 | 0.209 |
| 80% | 0.218 | 0.220 | 0.218 |
| 50% | 0.226 | 0.227 | 0.223 |
| 30% | 0.226 | 0.224 | 0.223 |
| 20% | 0.224 | 0.221 | 0.224 |
| 10% | 0.224 | 0.223 | 0.223 |
| 5% | 0.234 | 0.239 | 0.238 |
| 2% | 0.235 | 0.237 | 0.238 |
| 1% | 0.236 | 0.236 | 0.239 |
| 0% (Zero-shot) | 0.240 | 0.244 | 0.246 |

**Supplementary Table 5.** Full results (MSE and MAE) of UDE and baselines on Beijing climate data (ERA5)

| | MSE | MAE |
|---|---|---|
| ours (zero-shot) | 0.422 | 0.452 |
| ours (finetune) | 0.407 | 0.447 |
| PatchTST | 0.413 | 0.449 |
| iTransformer | 0.420 | 0.454 |
| DeepAR | 0.462 | 0.495 |
| Wavenet | 0.488 | 0.510 |
| SVR | 0.985 | 0.830 |
| ARIMA | 1.170 | 0.891 |

**Supplementary Table 6.** Full results (MSE and MAE) of UDE and baselines on Shanghai climate data (ERA5)

| | MSE | MAE |
|---|---|---|
| ours (zero-shot) | 0.445 | 0.472 |
| ours (finetune) | 0.386 | 0.440 |
| PatchTST | 0.397 | 0.441 |
| iTransformer | 0.417 | 0.454 |
| DeepAR | 0.455 | 0.498 |
| Wavenet | 0.497 | 0.524 |
| SVR | 0.986 | 0.834 |
| ARIMA | 1.041 | 0.851 |

**Supplementary Table 7.** Full results (MSE and MAE) of UDE and baselines on Guangzhou climate data (ERA5)

| | MSE | MAE |
|---|---|---|
| ours (zero-shot) | 0.438 | 0.466 |
| ours (finetune) | 0.370 | 0.425 |
| PatchTST | 0.390 | 0.434 |
| iTransformer | 0.386 | 0.431 |
| DeepAR | 0.459 | 0.497 |
| Wavenet | 0.494 | 0.517 |
| SVR | 0.958 | 0.804 |
| ARIMA | 1.088 | 0.822 |

**Supplementary Table 8.** Full results (MSE and MAE) of UDE and baselines on Tokyo climate data (ERA5)

| | MSE | MAE |
|---|---|---|
| ours (zero-shot) | 0.627 | 0.566 |
| ours (finetune) | 0.566 | 0.536 |
| PatchTST | 0.620 | 0.555 |
| iTransformer | 0.638 | 0.563 |
| DeepAR | 0.641 | 0.587 |
| Wavenet | 0.678 | 0.608 |
| SVR | 1.028 | 0.821 |
| ARIMA | 1.209 | 0.884 |

**Supplementary Table 9.** Full results (MSE and MAE) of UDE and baselines on London climate data (ERA5)

| | MSE | MAE |
|---:|---:|---:|
| ours (zero-shot) | 0.683 | 0.620 |
| ours (finetune) | 0.571 | 0.566 |
| PatchTST | 0.632 | 0.588 |
| iTransformer | 0.669 | 0.607 |
| DeepAR | 0.740 | 0.670 |
| Wavenet | 0.769 | 0.684 |
| SVR | 0.970 | 0.789 |
| ARIMA | 1.297 | 0.933 |

**Supplementary Table 10.** Full results (MSE and MAE) of UDE and baselines on Los Angeles climate data (ERA5)

| | MSE | MAE |
|---:|---:|---:|
| ours (zero-shot) | 0.601 | 0.550 |
| ours (finetune) | 0.479 | 0.489 |
| PatchTST | 0.531 | 0.507 |
| iTransformer | 0.532 | 0.505 |
| DeepAR | 0.575 | 0.544 |
| Wavenet | 0.620 | 0.566 |
| SVR | 1.087 | 0.815 |
| ARIMA | 1.291 | 0.862 |

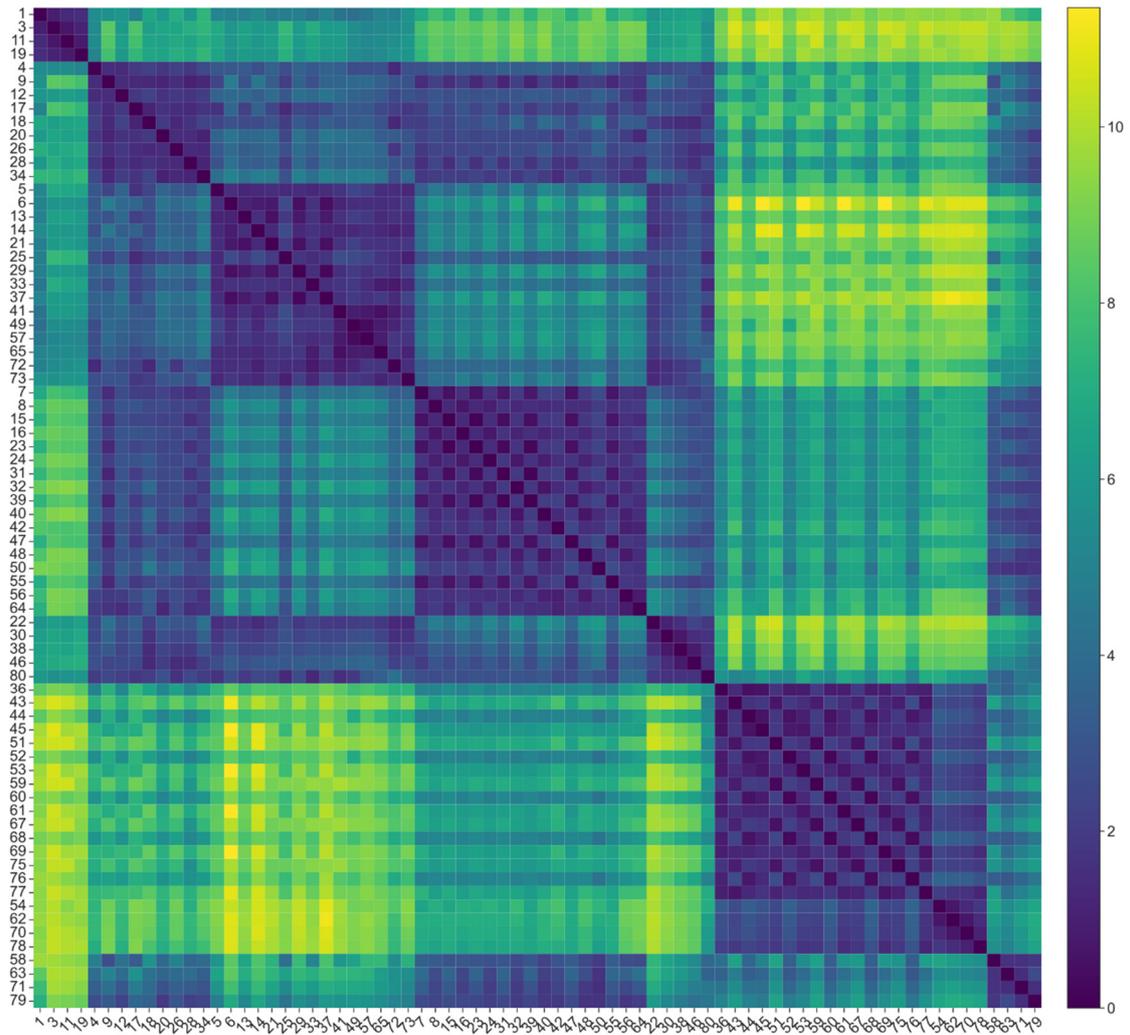

**Supplementary Fig. 1.** Full 2-Wasserstein distance matrix among all 81 subspace tokens constructed via time-delay patching (corresponding to Fig. 3c). Each token is derived from a local 2D patch of the delay-embedding matrix of a 512-step input sequence (ETTh1), and its topological structure is quantified via persistent homology on 1-dimensional features (loops). The 2-Wasserstein distance is computed between each pair of persistence diagrams, resulting in an 81 × 81 symmetric matrix. Distinct block structures reveal the presence of topologically coherent token clusters. The four representative tokens (Token 1, 3, 11, 19) forming the most prominent cluster—referred to as Pattern 1 in Fig. 3d—correspond to globally informative dynamical subspaces consistently highlighted in cross-layer attention maps (Fig. 3e–f).

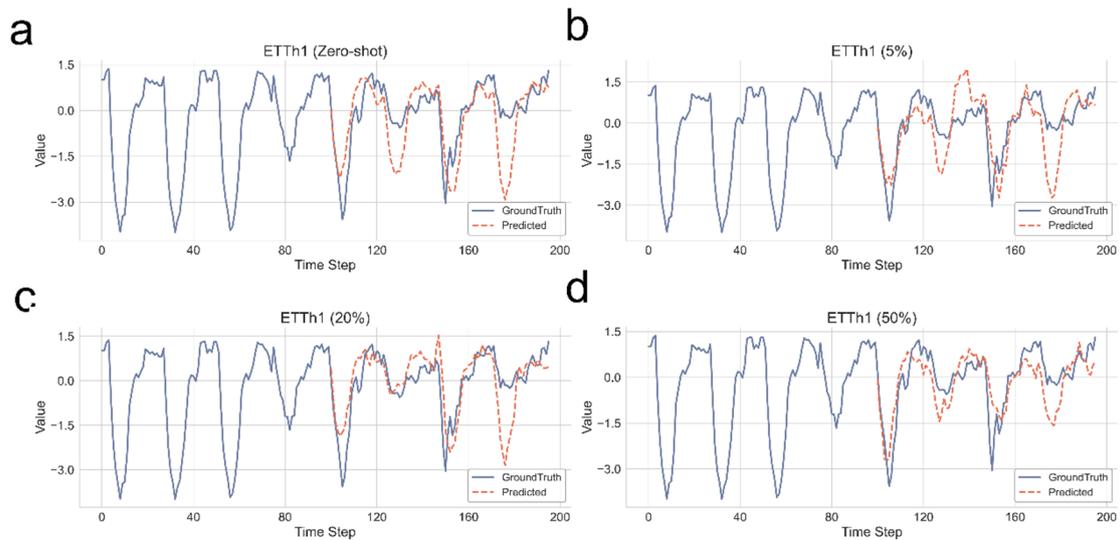

**Supplementary Fig. 2** Showcase of UDE forecasting performance on the ETTh1 dataset under different fine-tuning scenarios. Panels (a–d) illustrate model predictions (red dashed line) compared against the ground truth (blue solid line) over a 96-step forecasting horizon, given varying proportions of fine-tuning data: (a) Zero-shot (no fine-tuning), (b) 5%, (c) 20%, and (d) 50% of the training set. The results demonstrate progressive improvement in forecasting accuracy as the proportion of fine-tuning data increases, highlighting the data-efficient adaptation capability of the pretrained UDE model.

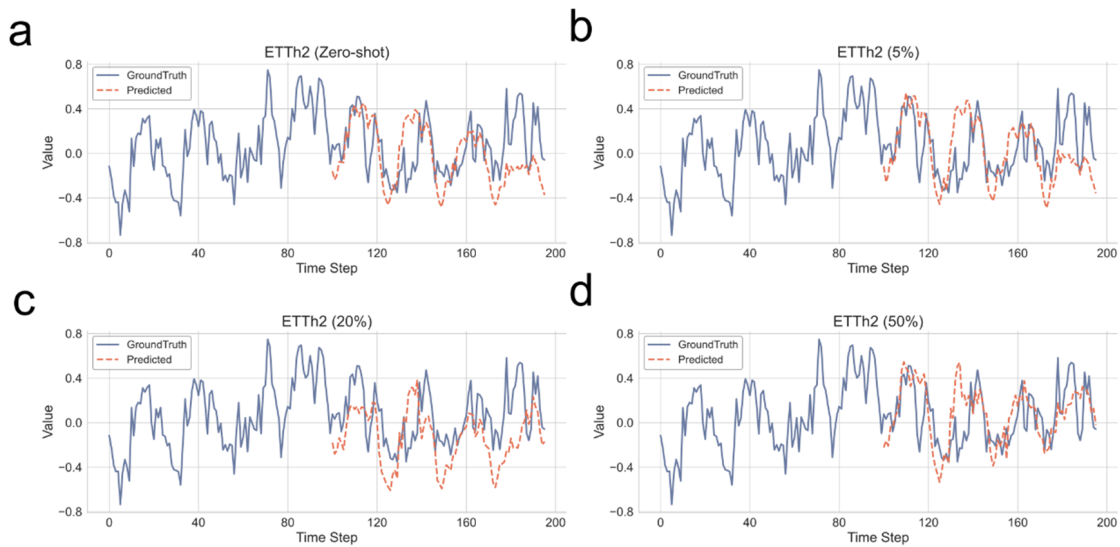

**Supplementary Fig. 3** Showcase of UDE forecasting performance on the ETTh2 dataset under different fine-tuning scenarios. Panels (a–d) illustrate model predictions (red dashed line) compared against the ground truth (blue solid line) over a 96-step forecasting horizon, given varying proportions of fine-tuning data: (a) Zero-shot (no fine-tuning), (b) 5%, (c) 20%, and (d) 50% of the training set. The results demonstrate progressive improvement in forecasting accuracy as the proportion of fine-tuning data increases, highlighting the data-efficient adaptation capability of the pretrained UDE model.

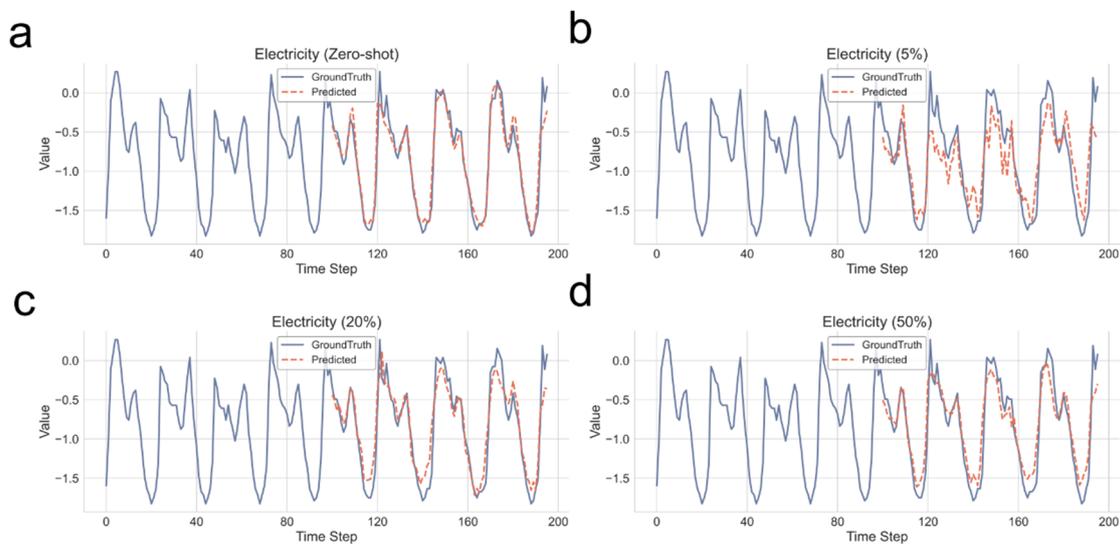

**Supplementary Fig. 4** Showcase of UDE forecasting performance on the Electricity dataset under different fine-tuning scenarios. Panels (a–d) illustrate model predictions (red dashed line) compared against the ground truth (blue solid line) over a 96-step forecasting horizon, given varying proportions of fine-tuning data: (a) Zero-shot (no fine-tuning), (b) 5%, (c) 20%, and (d) 50% of the training set. The results demonstrate progressive improvement in forecasting accuracy as the proportion of fine-tuning data increases, highlighting the data-efficient adaptation capability of the pretrained UDE model.

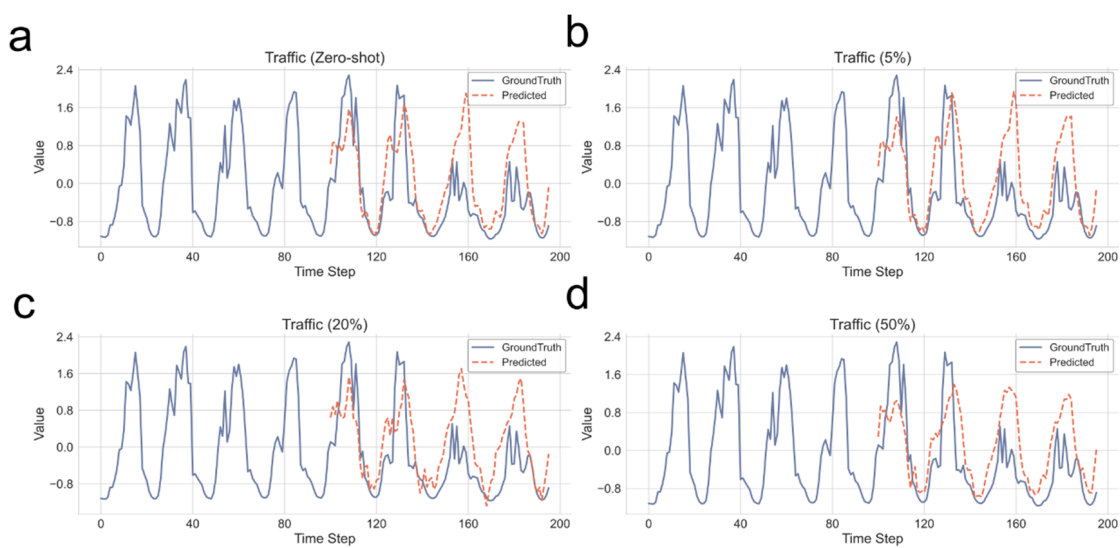

**Supplementary Fig. 5** Showcase of UDE forecasting performance on the Traffic dataset under different fine-tuning scenarios. Panels (a–d) illustrate model predictions (red dashed line) compared against the ground truth (blue solid line) over a 96-step forecasting horizon, given varying

proportions of fine-tuning data: (a) Zero-shot (no fine-tuning), (b) 5%, (c) 20%, and (d) 50% of the training set. The results demonstrate progressive improvement in forecasting accuracy as the proportion of fine-tuning data increases, highlighting the data-efficient adaptation capability of the pretrained UDE model.